\documentclass[a4paper]{article}
\pdfoutput=1
\usepackage[T1]{fontenc}
\usepackage[utf8]{inputenc}
\usepackage[USenglish]{babel}
\usepackage{float}
\usepackage{xcolor}
\usepackage{pdfcolmk}
\PassOptionsToPackage{normalem}{ulem}
\usepackage{ulem}
\usepackage{graphicx}
\usepackage[numbib,nottoc]{tocbibind}
\usepackage{cite}
\usepackage{amsmath,amsfonts,amsthm,amssymb,bm, bbm} % Math packages

\DeclareMathOperator*{\argmin}{arg\,min}
\usepackage[section]{placeins}
\usepackage{authblk}
\newcommand{\R}{\mathbb{R}}
\makeatother
\providecommand{\tabularnewline}{\\}
\floatstyle{ruled}
\usepackage{dirtree}
\usepackage{algorithm}
\usepackage{hyperref}
\makeatletter
\renewcommand{\ALG@name}{List}
\makeatother
\begin{document}
\author[1,2]{\small Ahmed Allam\thanks{ahmed.allam@uzh.ch}}
\author[3]{\small Mate Nagy\thanks{mate.nagy@yale.edu}} 
\author[5]{\small George Thoma\thanks{gthoma@mail.nih.gov}} 
\author[1,2,3,4]{\small Michael Krauthammer\thanks{michael.krauthammer@uzh.ch}}
\affil[1]{\footnotesize Department of Quantitative Biomedicine, University of Zurich}
\affil[2]{\footnotesize Biomedical Informatics Group, University Hospital of Zurich}
\affil[3]{\footnotesize Program in Computational Biology and Bioinformatics, Yale University}
\affil[4]{\footnotesize Department of Pathology, Yale School of Medicine}
\affil[5]{\footnotesize Lister Hill National Center for Biomedical Communications, National Library of Medicine}
\title{Neural networks versus Logistic regression for 30 days
all-cause readmission prediction}
\date{\vspace{-5ex}}
\maketitle
\begin{abstract}
Heart failure (HF) is one of the leading causes of hospital admissions
in the US. Readmission within 30 days after a HF hospitalization is
both a recognized indicator for disease progression and a source of
considerable financial burden to the healthcare system. Consequently,
the identification of patients at risk for readmission is a key step
in improving disease management and patient outcome. In this work,
we used a large administrative claims dataset to (1) explore the systematic
application of neural network-based models versus logistic regression
for predicting 30 days all-cause readmission after discharge from
a HF admission, and (2) to examine the additive value of patients'
hospitalization timelines on prediction performance. Based on data
from 272,778 (49\% female) patients with a mean (SD) age of 73 years
(14) and 343,328 HF admissions (67\% of total admissions), we trained
and tested our predictive readmission models following a stratified
5-fold cross-validation scheme. Among the deep learning approaches,
a recurrent neural network (RNN) combined with conditional random
fields (CRF) model (RNNCRF) achieved the best performance in readmission
prediction with 0.642 AUC (95\% CI, 0.640-0.645). Other models, such
as those based on RNN, convolutional neural networks and CRF alone
had lower performance, with a non-timeline based model (MLP) performing
worst. A competitive model based on logistic regression with LASSO
achieved a performance of 0.643 AUC (95\% CI, 0.640-0.646). We conclude
that data from patient timelines improve 30 day readmission prediction
for neural network-based models, that a logistic regression with LASSO
has equal performance to the best neural network model and that the
use of administrative data result in competitive performance compared
to published approaches based on richer clinical datasets.
\end{abstract}
\newpage{}
\section{Introduction}

Heart failure (HF) is one of the leading causes for hospital admissions
in the US \cite{Fingar2015,Fingar2017,Bergethon2016,Desai2012} with
high numbers of readmissions within 30 days of discharge \cite{Desai2012,Fingar2017,Bergethon2016}.
Based on multiple hospitalization data sources, the yearly rate of
30 days all-cause readmission after an HF hospitalization is approximately
23-24\% \cite{Fingar2015,Fingar2017,Ross2009}, posing a huge burden
on the healthcare system with an estimated cost of \$17 billions of
total Medicare expenditure \cite{Desai2012,Arundel2016}. Beyond the
associated expenses and costs, readmissions have negative consequences
on patients\textquoteright{} health status, leading to complications
and increased risk of disease progression \cite{Arundel2016}. Efforts
toward quality improvement such as introducing programs that incentivize
and penalize hospitals based on the yearly readmission rate have been
the focus of researchers and policy makers \cite{Fingar2017,Ziaeian2016}.
Likewise, there has been increasing interest in developing predictive
models and/or monitoring systems that allow for prevention and preemptive
steps, such as the prediction of 30 days all-cause readmission for
patients hospitalized with HF for which many challenges remain \cite{Mortazavi2016,JD2017}.
\newline In this paper, we aim at exploring the systematic application
of neural network models for predicting 30 days all-cause readmission
after discharge from a HF hospitalization (which we call index event
below). Concretely, given a set of sequences of hospitalization admissions
with their corresponding 30 days all-cause readmission outcome, we
seek to predict the 30 days all-cause readmission of the last HF admission
(i.e. the last index event) in each sequence. The sequence of hospitalization
events for each patient will be referred to as ``\textit{timeline}''
and ``\textit{trajectory}'' interchangeably throughout the paper.
Published approaches chiefly use data from the index event for predicting
hospital readmission, paying less attention to a patient's trajectory
leading to the current heart failure admission. Intuitively, a patient's
history may add much additional information that may be informative
of whether a patient is subject to early readmission. For example,
a history of multiple readmissions in the past may be a risk factor
for future readmissions. Consequently, one specific aim of this study
is to examine the value of including a patient's trajectory data in
a 30 day readmission prediction model. To this end, we examine three
approaches for modeling the problem of which two use the temporal
information encoded in the patients' trajectories (sequence labeling
and sequence classification), and one that does not (index event
classification). Particularly, we implemented multiple neural network
models with varying architectures and objective functions such as
recurrent neural networks (RNN), and convolutional neural networks
(CNN) as examples of sequence labeling and classification approaches,
and multilayer perceptron (MLP) along with logistic regression as
baseline models representing the index event classification approach.
We conducted these studies with a large administrative claims dataset,
which lacks the detailed clinical information found in datasets typically
used for this problem. As claims data are readily available and can
be robustly harmonized, they pose less privacy concerns and are ideally
suited for tacking the HF readmission problem.

\section{Methods}

\subsection{Dataset}

The HF dataset was derived from the Healthcare Cost and Utilization
Project (HCUP), Nationwide Readmission Database (NRD), issued by the
Agency for Healthcare Research and Quality (AHRQ) \cite{HealthcareCostandUtilizationprojectHCUP.AgencyforHealthcareResearchandQuality}.
It includes patients' discharges (i.e. hospital claims) of all-payer
hospital inpatient stays over the 2013 period that are contributed
by twenty one states and accounting for 49.1\% of all US hospitalizations
\cite{HealthcareCostandUtilizationprojectHCUP.AgencyforHealthcareResearchandQuality}.
Each claim in the dataset is associated with a corresponding patient
who is identified by a uniquely generated linkage number (``\textit{\small{}visitlink}'')
that tracks the patient's visits across hospitals within a state.
Each claim represents a summary of an inpatients hospitalization event,
including information about the hospitalization event such as the
time of admission and discharge, the diagnosis, procedures, comorbidity
and chronic conditions, length of stay, along other clinical fields
associated with the event (a detailed description of the data elements
can be found at\cite{HCUPNationwideReadmissionDatabasetNRD}). Moreover,
as each claim is linked to a patient identifier, it also includes
patient's socio-demographic information such as age, gender, income
category and place location based on the National Center for Health
Statistics (NCHS) classification scheme for US counties. 

\subsubsection{Timeline/trajectory building and processing}

We built timelines/sequences out of the claims, allowing us to preserve
the temporal progression and the history of hospitalization events
for every patient. Patients were included in the HF dataset if they
met the following conditions:
\begin{enumerate}
\item had at least one hospitalization event between January and November
period with HF as the primary diagnosis (i.e. congestive heart failure;
code = 108) as determined by Clinical Classification Software (CCS)
that groups International Classification of Diseases, Version 9 (ICD-9)
codes \cite{AgencyforHealthcareResearchandQuality2009} 
\item were $\geq$ 18 years old when they had an HF hospitalization event
\end{enumerate}
Formally, we denote each claim (i.e. hospitalization event) by a feature
vector $\overline{x}_{t}$ describing the characteristics and attributes
of the hospitalization event and the corresponding patient. Moreover,
we denote its corresponding label by $y_{t}\in\{0,1\}$, representing
the 30 days all-cause readmission . The readmission outcome was computed
based on the AHRQ HCUP 30-day readmission measure (see Appendix A
in \cite{Barrett2012}).

\begin{figure}
\caption{An example of a patient's timeline with 30 days all-cause readmission
labeling}

\includegraphics[scale=0.37]{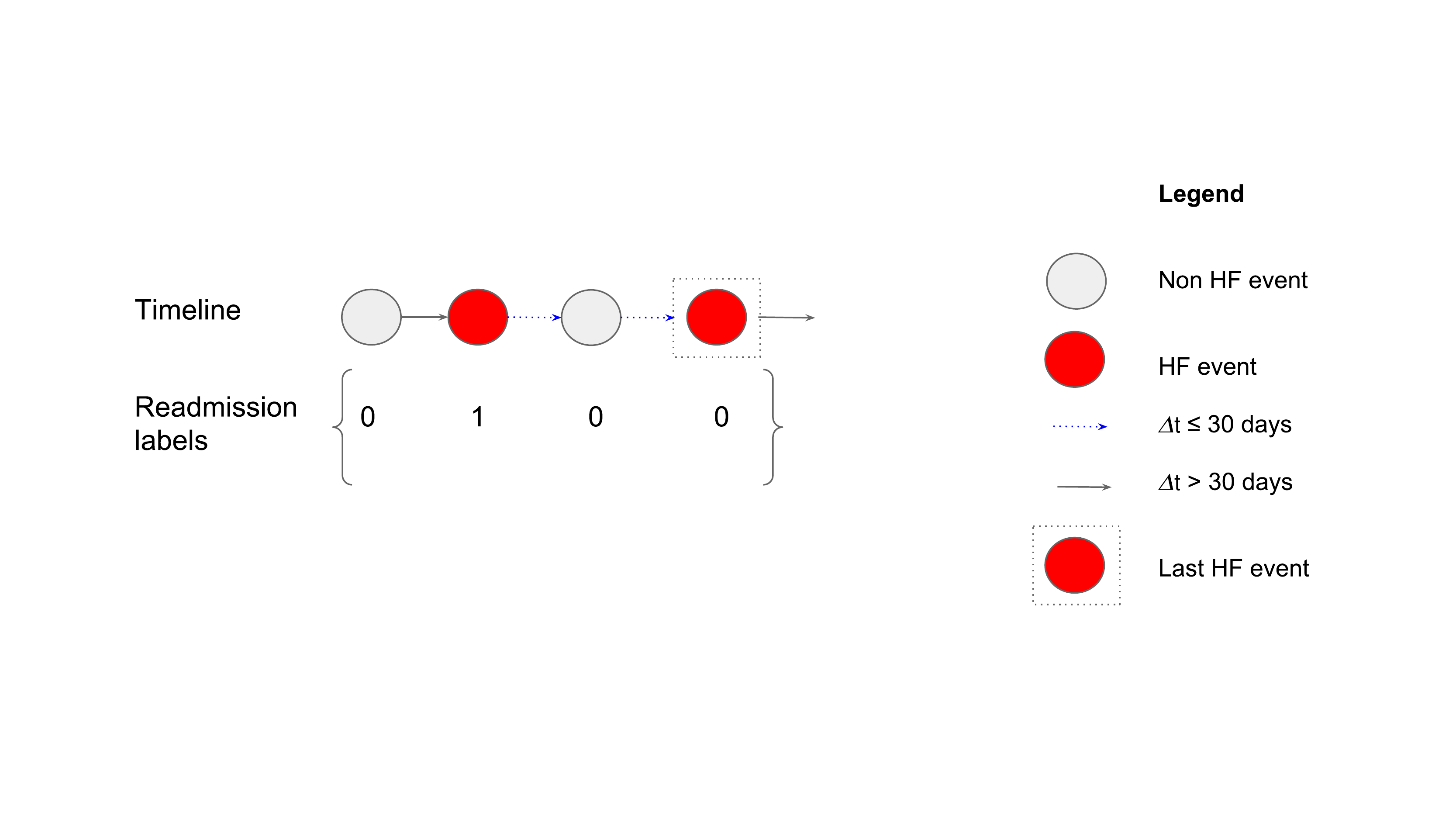}

\label{fig:timeline_patient}
\end{figure}

To determine if $y_{t}=1$ (i.e. the hospital admission of the future
claim/event $\overline{x}_{t+1}$ occurs within 30 days from the current
event $\overline{{x}}_{t}$), we traverse the patient's timeline (temporally-ordered
hospitalization events) from left to right and check if:
\begin{enumerate}
\item the current event $\overline{x}_{t}$ is an index event (i.e. an event
where HF is the primary diagnosis as indicated by CCS diagnosis grouper;
code = 108) and
\item the difference between the admission of the next event $\overline{x}_{t+1}$
and the discharge of current event $\overline{x}_{t}$ is $\leq$
30 days (i.e. $\Delta t\leq30$ days)
\end{enumerate}
Figure \ref{fig:timeline_patient} depicts the labeling process of a
patient's timeline. Notice the final event will always be the last
HF event in a patient's timeline for which we can determine its readmission
label.

\subsubsection{Dataset features}

Each claim/event in a patient's timeline was represented by a feature
vector $\overline{x}_{t}$ encoding the characteristics of the hospitalization
event and the corresponding patient. The feature vector included most
of the fields included in the NRD databases describing every inpatients
hospitalization event such as the time of admission and discharge,
the diagnosis, procedures, comorbidity and chronic conditions, length
of stay, along other clinical fields associated with the event. A
detailed description of all the used features is found in the supplementary
material.

\subsection{Models and notation}

\subsubsection{Sequence labeling and classification\label{subsec:Sequence-labelling-preliminaries}}

In this section, we introduce the sequence labeling approach to 30
days all-cause readmission prediction. Generally, given a patient's
temporally ordered sequence of claims \sloppy $\mathbf{\underline{x}}=[\overline{x}_{1},...,\overline{x}_{t},...,\overline{x}_{T}]$,
represented by a $d$-dimensional feature vector $\overline{x}_{t}\in\R^{d}$,
we seek a labeling $\mathbf{\underline{y}}$ = $\ensuremath{[y_{1},...,y_{t},...,y_{T}]}$
representing the 30 days all-cause readmission outcomes where $y_{t}\in\{0,1\}$
and $T$ is the patient-specific sequence length (i.e. equivalent
to $T_{i}$ where $i$ refers to the $i$-th patient in training dataset).
Given a training set $D_{train}\ensuremath{=}\{(\underline{\mathbf{x}_{i}},\underline{\mathbf{y}_{i}})\}_{i=1}^{N}$,
the goal is to learn a model (i.e. function map $f$) by minimizing
an objective function $L(f,D_{train})$ that measures the discrepancy
between every sequence's target labels $\mathbf{\underline{y}}_{i}$
and its corresponding predicted label sequence $\mathbf{\underline{\hat{y}}}_{i}$
in the training dataset. A common approach is to use a parametrized
function $f(\bm{\theta})$ such that learning the best function map
(i.e. training a model) translates into finding the optimal weights
$\bm{\theta}$ where $\bm{\theta}=\argmin_{\bm{\theta}}L(f,D_{train})$.
With the choice of a differentiable  function, the optimal weights $\bm{\theta}$ are obtained
through an iterative process by using the gradient of the objective
function $\nabla_{\bm{\theta}}L(f,D_{train})$, scaling it with step
size $\eta$, and subtracting the result from the current weights
at each iteration. Intuitively, the weights update equation 
\begin{equation}
\ensuremath{\bm{\theta}^{k+1}=\bm{\theta}^{k}-\eta\nabla_{\bm{\theta}^{k}}L(f,D_{train})}\label{eq:vanilla_grad_descent}
\end{equation}
is directing the new weights toward the steepest descent (i.e. the
direction which minimizes $L(f,D_{train})$) at each update iteration
$k$. Sequence classification is similar to sequence labeling but
instead of assigning labels/classes to each event in the sequence,
one assigns one single label/class to the whole sequence.  Thus,
in the sequence labeling setting, the trained model will predict an
outcome for every event in the sequence, while in the sequence classification
setting, the model predicts the class of the whole sequence. In this
work, the difference between both approaches is mainly in the training
phase (learning the labels of all events versus one single label for
the sequence), while during the testing phase, both models are used
to predict the outcome/label of the last HF event. The latter is directly
provided through sequence labeling. In sequence classification, we
are using the label of the last HF event as a substitute for the sequence
label and a training loss that is associated with that event. To summarize,
we use the term ``labeling'' and ``classification'' to differentiate
between models incorporating the labels of previous events in the
training/learning of the model versus optimizing only on the last
HF event label. In all cases, the testing/decoding phase is equivalent
and is focused on the prediction of the label/class of the last HF
event.

\subsubsection{Recurrent neural network (RNN)}

Recurrent neural networks (RNN) is a connectionist model that is well
suited for modeling sequential and temporal data with varying length
\cite{Elman1990,Graves2012,Lipton2015a}. A basic RNN is similar to
feed-forward neural network but with additional support for cyclical
connections (i.e. recurrent edges among the hidden layers at different
time steps) \cite{Lipton2015a,Graves2012}. RNNs computes a hidden
vector at each time step (i.e. state vector $\overline{h}_{t}$ at
time $t$), representing a history or context summary of the sequence
using the input and hidden states vector form the previous time step.
This allows the model to learn long-range dependencies where the network
is unfolded as many times as the length of the sequence it is modeling.
To compute the outcome $\hat{y}_{t}$ at time $t$, an affine transformation
followed by non-linear activation function $\sigma$ is applied to
the state vector $\overline{h}_{t}$. The non-linear operator $\sigma$
can be either the $sigmoid$ function applied to a scalar input or
its generalization to multi-class $softmax$ function applied to
vector. As a result, the outcome $\hat{y}_{t}$ represents a probability
distribution over the set of possible labels at time $t$. Gradient
descent (i.e. ``\textit{vanilla}'' gradient descent as in Equation \ref{eq:vanilla_grad_descent} or any variant)
is used for optimizing the weights of the network while the gradient
is computed using back propagation through time\cite{Werbos1990}.
Although RNNs are capable of handling and representing variable-length
sequences, in practice, the learning process faces challenges due
to the vanishing/exploding gradient problem \cite{Hochreiter1991,Bengio1994,Graves2012}.
To overcome these challenges, gradient clipping \cite{Pascanu2013}
and gated memory cells approach as in long short-term memory (LSTM)
and gated recurrent unit (GRU) \cite{Hochreiter1997,Cho2014,chung2014empirical} were
proposed replacing the conventional nodes in the hidden layer and
hence updating the computation mechanism of the hidden state vector
$\overline{h}_{t}$.

\subsubsection{RNN objective function\label{subsec:RNN-objective-function}}

We defined the loss at each time step for an $i$-th sequence by
the cross-entropy error/loss

\begin{equation}
l_{t}^{(i)}=-\sum_{c=1}^{|V_{label}|}y_{t,c}^{(i)}\times log(\hat{y}_{t,c}^{(i)})
\end{equation}
where $V_{label}$ is set of admissible classes, $|V_{label}|$ is the number of classes, $y_{t,c}\in\{0,1\}$ is equivalent to $\mathbbm{1}{\big[y_{t} = c\big]}$ (i.e. a boolean indicator that is equal to 1 when $c$ is the reference/ground-truth class at time $t$),
and $\hat{y}_{t,c}$ is the probability of the class $c$
at time $t$. Four realizations/definitions of objective functions
were tested in this study. Given that our focus is on the 30 days
all-cause readmissions for the last HF hospitalization event, the
first loss (Convex\_HF\_lastHF) was defined by a convex combination
between the average loss from all HF events in patient's timeline
and the loss from the last HF event. The convex combination is parametrized
by parameter $\alpha$ that was determined using a validation set
inspired by the work done in \cite{Lipton2015}. The second loss function
(LastHF) used the loss computed only from the last HF event
while the third (Uniform\_HF) uniformly averaged the loss
from all HF events in a patient's timeline. Lastly, the forth objective
function (Convex\_HF\_NonHF) was based on a convex combination
between the average loss contributed by all HF events in patient's
timeline and the average loss from the non HF events.
\newline
The objective function for the whole training set $D_{train}$ was defined by the average loss $L_i$ across all
the sequences in $D_{train}$ plus a weight regularization term (i.e.
$l_{2}$-norm regularization) applied to the model parameters represented
by $\bm{\theta}$

\begin{align}
L_i & =\frac{1}{T_i}\sum_{t=1}^{T_i}l_{t}^{(i)} \\
L(\mathbf{\bm{\theta}}) & =\frac{1}{N}\sum_{i=1}^{N}L_{i}+\frac{\lambda}{2}||\mathbb{\bm{\theta}}||_{2}^{2}\label{eq:trainnn_obj}
\end{align}
\newline
In addition to the $l_{2}$-norm regularization in the objective function,
we also experimented with dropout \cite{Srivastava2014} by deactivating
neurons in the network layers using probability $p_{dropoout}$ in
order to reduce the network's chances of overfitting the training
set .

\subsubsection{RNN with scheduled sampling (RNNSS)}

Another variation of the RNN model that we experimented with is using
a scheduled sampling approach (``teacher forcing'') \cite{NIPS2015_5956}
while training the RNN model. Again we used the same four definitions
of the loss functions used in the RNN case. 

\subsubsection{Conditional random fields (CRF)}

Although RNN models are suited for modeling temporal data, the outcome/label
prediction for each event is preformed independently from each other.
 That is: the labeling decision is done \textit{locally} (i.e. without
considering any association/correlation between neighboring labels).
In other words, there is a need for a joint modeling approach that
is \textit{global} by considering the whole sequence of labels when
performing the optimization and inference. Linear-chain CRF suits
this requirement well by modeling the probability of the whole labeled
sequence (i.e. outcome sequence) given the input sequence. It is a
class of undirected discriminative graphical models that uses a global
feature function within a log-linear model formalism, making it well
suited for structured prediction\cite{Lafferty2001}. In this study,
we applied CRF in two occasions with two variations (i.e. definition
of potential functions).

\subsubsection{CRF with RNN}

We first experimented with combining the RNN model with a CRF layer
by feeding the computed features from the RNN layer as inputs to the
CRF layer as in \cite{Lample2016}. We denote the output features
of the RNN layer by $\mathbf{\underline{z}}=[\overline{z}_{1},\overline{z}_{2},\cdots,\overline{z}_{T}]$
representing the sequence of output features computed from the input
sequence $\mathbf{\underline{x}}$ (both sequences have equal length).
The potential functions in the CRF layer were computed using $\mathbf{\underline{z}}$
along with label sequence $\mathbf{\underline{y}}$ in two variations: 
\begin{enumerate}
\item RNNCRF (Unary) that computed unary potential by using only the RNN
output feature vector to generate an output vector with dimension
equal to the number of classes $|V_{label}|$ for each $\overline{z}_{t}$.
The pairwise potential is modeled using a transition parameters matrix
$A(y_{t-1},y_{t})$ of size $|V_{label}|\times|V_{label}|$ representing
the transition score from one outcome class to another. 
\item RNNCRF (Pairwise) that computes pairwise potentials using both the
RNN output feature vectors and the labels sequence such that it generates
an output vector of size $|V_{label}|\times|V_{label}|$ at every
time step $t$ similar to the approach reported in \cite{Ma2016}. 
\end{enumerate}

\subsubsection{CRF \& Neural CRF}

We also tested a CRF approach without the RNN block. The first used
CRF only (i.e. first-order linear chain CRF) model using the two variations
of potential functions (i.e. unary and pairwise). The second model
is combining CRF with neural model (i.e. using non-linear transformation
for computing features) similar to the approach in\cite{Peng2009} using
the same two potential function variants.$\newline$The objective
function for models that incorporated CRF was defined by the negative
conditional log-likelihood $L(\bm{\theta})$ plus an $l_{2}$-norm
weight regularization term,

\begin{equation}
L(\bm{\theta})=\left[\frac{1}{N}\sum_{i=1}^{N}-log(p(\mathbf{\underline{y_{i}}}|\mathbf{\underline{x_{i}}}))\right]+\frac{\lambda}{2}||\mathbb{\bm{\theta}}||_{2}^{2}\label{eq:crf_objfunc}
\end{equation}
Estimating the optimal weights $\bm{\theta}$ is typically done by
applying a variant of gradient descent algorithm (as described in
Equation \ref{eq:vanilla_grad_descent}) where the sum-product algorithm
(i.e. performing a variation of the forward-backward algorithm \cite{Bishop2006})
is used. Decoding the sequence (i.e. finding the optimal labeling
$\hat{\mathbf{\underline{y}}}_{optimal}$) is done through a variant
of Viterbi algorithm \cite{Viterbi1967,Collins}. 

\subsubsection{Convolutional neural networks (CNN)}

The CNN models adopt the sequence classification view by using the
2$D$ arrangement of the patients' timelines with an objective function
defined only for the last HF event (i.e. the loss function is defined
for the last HF event that we seek to predict its readmission outcome).
A CNN model is a feed-forward neural network that typically consists
of multiple layers of which \textit{convolutional} layer is the building
block. A convolutional layer is composed of filters/kernels (in our
context, the kernel is a 2$D$ arrangement of weights in matrix form)
that are convolved with the features of the previous layer (such as
the input layer) to produce feature maps. More formally, a patient's
timeline was arranged in a matrix form where the sequence of events
are stacked (i.e. concatenated) to form a matrix $X=\begin{bmatrix}\overline{x}_{1} & \overline{x}_{2} & \cdots & \overline{x}_{T_{max}}\end{bmatrix}^{\intercal}$
of size $T_{max}\times d$ where $d$ is the dimension of an event
vector $\overline{x}_{t}$ and $T_{max}$ is the maximum length of
a patient's timeline in the training set \textendash{} patients with
shorter timelines are padded to have a common representation. A kernel
$F$ is a matrix of weights that is convolved with $X$ to produce
a feature map $M$ such that an entry in $M$ is computed by first
taking the sum of element-wise multiplication of the weights in the
kernel $F$ and the corresponding input of the previous layer, then
adding a bias term followed by non-linear operation. Typically, multiple
kernels are applied and the resulting feature maps are stacked on
top of each other forming a 3$D$ volume/tensor to be processed subsequently
in the next layers. The elements of each kernel represent the shared
parameters that we optimize during the training phase.
Another type of layers in this network is a \textit{pooling} layer
that also includes kernels/filters but with no trainable weights,
which slides over the input feature maps based on a defined horizontal
and vertical stride size and computes a summary score such as a maximum
or average score for every region of overlap. As a result, in the
pooling layer we can change the size of the generated feature maps
by specifying the stride and padding size such that the size of the
feature maps decreases as we progress into subsequent layers in the
network (i.e. equivalent to subsampling). Another commonly used layer
after the convolutional/pooling layers is the \textit{fully-connected}
layer (FC). FC takes an input vector from the reshaped feature maps
generated in the last convolutional/pooling layers and applies an
affine transformation followed by non-linear element-wise operation.
In this work, we experimented with two types of convolutional models: 
\begin{enumerate}
\item CNN model that describes a network inspired by commonly used models
in computer vision and image processing research \cite{Simonyan2015}
that makes use of multiple small square convolutional and pooling
kernels, where the generated feature maps are reduced in size as a
function of the network depth (i.e. number of layers) until reaching
to the fully-connected layer/s.
\item CNN-Wide model that adapts the approach used by Kim \cite{Kim2014}
for sentence classification where the convolutional kernels are wide/rectangular
covering the whole input feature dimension. In other words, a kernel
in this model would have varying sizes (such as $2\times d$, $3\times d$,
$5\times d$ ) where the convolution is applied to the whole feature
vector for two or more events for every possible window of events
in the patient's timeline (i.e. applied to matrix $X$). After each
convolution operation, the result is a vector of feature map for every
kernel. In this network, the pooling layer reduces each generated
feature map vector to a scalar (i.e. one feature) and then concatenates
each one of them into one vector having number of elements equal to
the number of applied convolutional kernels. Lastly, the resulting
vector is passed into one or more FC layers before it is passed to
the output layer.
\end{enumerate}
Both CNN models use an output layer where the computed vector of activations/feature
map in the penultimate layer are passed to generate a probability
distribution over the outcome labels (as in the RNN case). The defined
loss function for every sequence in both models is the loss computed
for the last HF event as in the RNN case (see LastHF in Section
\ref{subsec:RNN-objective-function}) where the overall objective
function for the training set is also equivalent to RNN case (Equation
\ref{eq:trainnn_obj}). 

\subsubsection{Multilayer perceptron (MLP)}

A final neural network-based model is the multilayer perceptrons which
is also a feed-forward neural network (MLP). The MLP network is composed
of an input layer then a set of multiple FC layers and lastly an output
layer that generates a probability distribution over the outcome classes.
The FC layers, as we discussed earlier, mainly consists of two operations;
an affine transformation followed by non-linear element-wise operation
to generate new feature vectors (i.e. learned representations). The
difference between this modeling approach and the previous ones is
that MLP takes the \textit{event} view of the problem by modeling
the last index event $\overline{x}_{T}$ only and discarding the sequence
aspect of the patients' timeline. The defined loss and the overall
objective function is equivalent to the ones defined for the RNN case
(LastHF) and Equation \ref{eq:trainnn_obj}.

\subsubsection{Logistic regression (LR)}

Logistic regression (LR) is a commonly used model for classification
problems due to its simplicity and model interpretability. Like MLP,
LR supports the \textit{event} view of the problem by modeling only
the last index event. LR model can be considered as a neural network
model with no hidden layers and one output neuron. In this setup,
the input features are fully-connected to one output neuron where
the $sigmoid$ function is applied as a non-linear operation computing
the probability of the outcome label to be equal to 1. In other words,
the LR model computes $p(\hat{y}_{T}=1|\overline{x}_{T})=\frac{1}{1+\exp^{\mathbf{-(W_{1\times d}\overline{x}_{T}+b)}}}$
where $W_{1\times d}$ is the weight matrix that maps $\overline{x}_{T}$
to a scalar value (i.e. using one neuron), $b$ is the bias term and
$\frac{1}{1+\exp^{-z}}$ is the $sigmoid$ function representing the
non-linear operation. The output represents the probability of a patient
readmitting to hospital within 30 days after HF hospitalization event.
In this work, LR was the baseline model that we compare its performance
to the ones of the neural network-based models. The loss function
for each patient's last event is defined by the conditional log-likelihood
which is equivalent to the cross-entropy loss for the binary case
(i.e 2-class classification) and the overall objective function is
based on the average conditional log-likelihood of the data (see Equation
\ref{eq:trainnn_obj}). Additionally, we experimented with two regularization
schemes: (1) $l_{1}$-norm regularization (LASSO) and (2) $l_{2}$-norm
regularization.

\section{Experimental setup}

We followed a stratified 5-fold cross-validation scheme, in which the
HF dataset is split into 5 folds, each having a training and test set
size of 80\% and 20\% of the data, respectively, and a validation set size of 10\% of
the training set in each fold (used for optimal epoch selection in case of neural models or hyperparameter selection in case of logistic regression). Moreover, due to the imbalance in outcome classes (i.e. no readmission vs. readmission), training examples were weighted inversely proportional to class/outcome frequencies in the training data. The models' performance was evaluated
using the last HF event in the patients' timeline (i.e. 30 days all-cause readmission after hospitalization for HF event). We used the area
under the ROC curve (AUC) as our performance measure with confidence
intervals computed using the approach reported in LeDell et al. \cite{LeDell2015}. Moreover, the evaluation of the trained models was based on their
average performance on the test sets of the five folds. 

\subsection{Hyperparameter optimization for neural models}

Neural model hyperparameter selection is costly, particularly for finding the optimal architecture. To this end, we randomly chose one fold
where 30\% of the training set was further split into a training and
validation set, each having 90\% and 10\% of the data, respectively. We developed a multiprocessing
module that used a uniform random search strategy \cite{BergstraJAMESBERGSTRA2012}
that randomly chose a set of hyperparameters configurations (i.e.
layer depth, filter size and optimization methods, see supplementary materials for more details) from the set of all possible
configurations.
Then the best configuration for each model (i.e. the one achieving
best performance on the validation set) was used for the final training
and testing. 

\section{Results}

\begin{figure}
\includegraphics[scale=0.3]{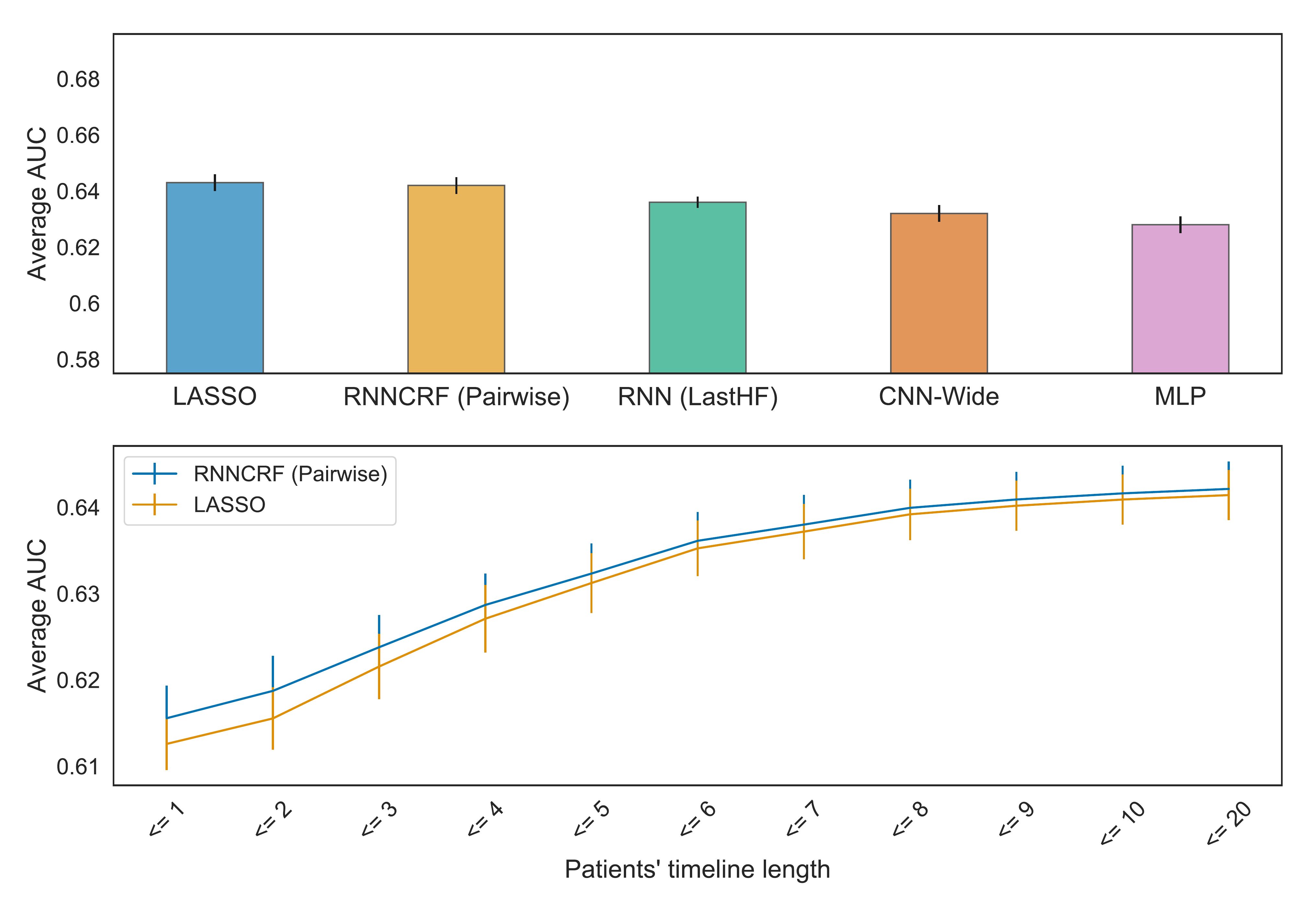}

\caption{\label{fig:Ranking-of-best}Ranking of best models (upper panel).
Performance of LASSO versus RNNCRF model as function of patients'
timeline length (lower panel)}
\end{figure}

\begin{figure}
\includegraphics[scale=0.25]{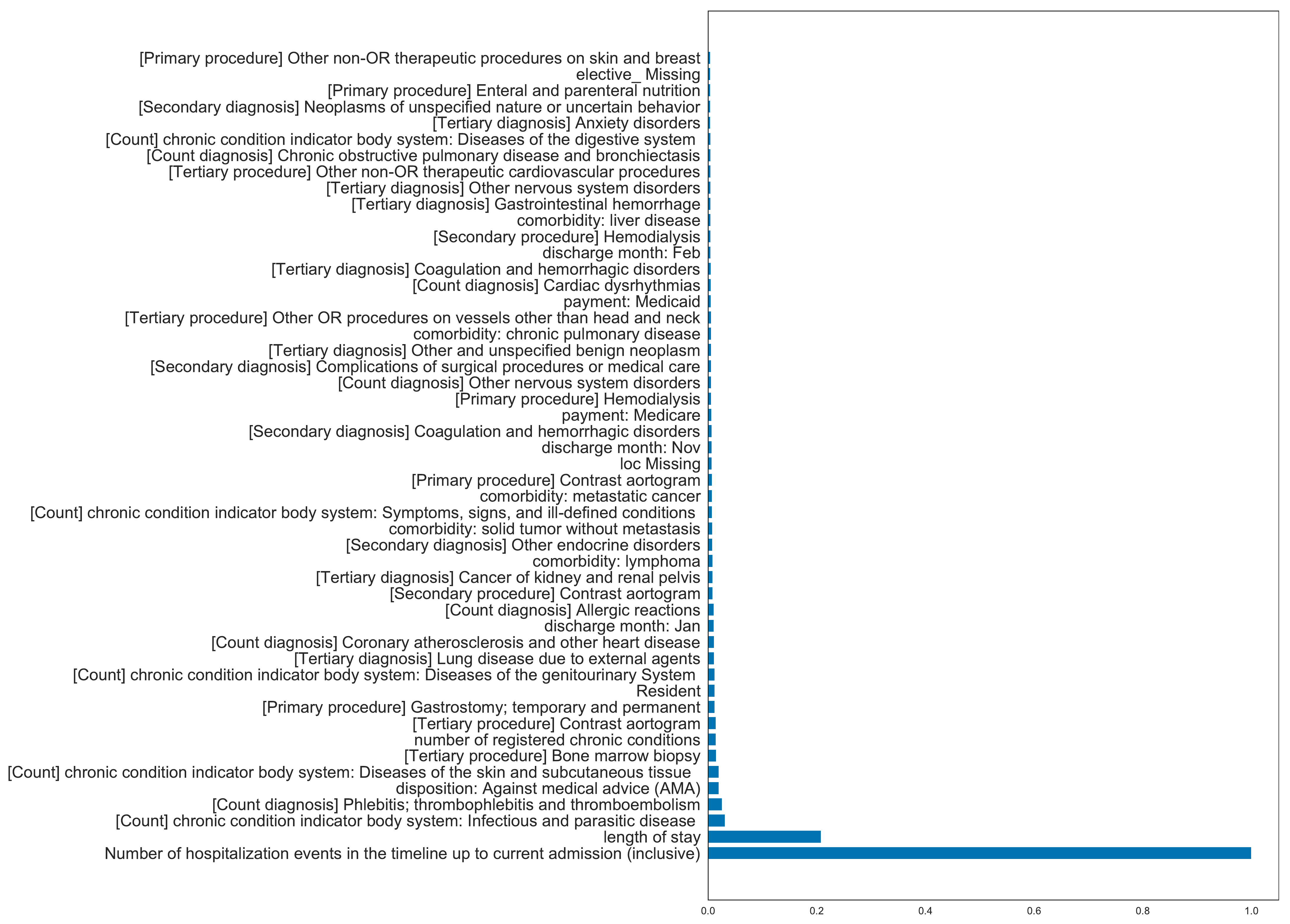}

\caption{\label{fig:Top-50-features-in}Top-50 features in LASSO models contributing to the increase of log-odds of readmission}
\end{figure}

The HF dataset included 272,778 patients (49\% female) with a mean
(SD) age of 72.89 years (14). The total number of HF admissions was
343,328 (66.9\% of all admissions) of which 81,087 (23.6\%) were
30 days all-cause readmissions, corresponding to the official rates
published by HCUP\cite{Fingar2017}. Among the last HF hospitalizations in
patients' timelines, 45,183 (16.6\%) resulted in readmissions. Table
\ref{tab:Overview-of-HF} reports a general overview of the characteristics
of the dataset including socio-demographics, hospitalization events,
top diagnosis and procedures and the payment source. Table \ref{tab:AUC-models'-performance}
reports the models' performance in predicting the 30 days all-cause
readmission for the last HF event in every patient's timeline. Starting
from RNN, the models trained with losses incorporating/emphasizing
the loss from last HF event (i.e LastHF and Convex\_HF\_LastHF) achieved
higher performance 0.636 and 0.635 AUC respectively compared to other
loss definitions. Moreover, RNN models with all four loss definitions
achieved higher performance than RNNSS counterparts. For models incorporating
CRF, the RNNCRF model achieved the highest performance with 0.642,
followed by Neural CRF 0.634 and CRF only model achieving 0.63 with
the first two models using pairwise potentials and the last one using
unary potential. For convolutional models, CNN-Wide achieved better
performance 0.632 compared to conventional CNN with 0.619. The MLP
model achieved 0.628 placing it as the lowest performing model among
the classes of neural models (see Figure \ref{fig:Ranking-of-best} top panel).
The baseline model LR with $l_{1}$-norm regularization (LASSO) achieved
higher performance 0.643 compared to LR with $l_{2}$-norm regularization
0.637. The ranking of the best performing models is depicted in Figure
\ref{fig:Ranking-of-best} where the bottom panel compares the performance of the LASSO to the RNNCRF model as a function
of the length of patients' timeline. As the length of the timeline
increases (i.e. more hospitalization events), the gap in last HF event prediction
performance between both models decreases. The analysis of feature importance is
reported in Figure \ref{fig:Top-50-features-in}, which shows the normalized
coefficients of the trained LASSO models averaged across all folds.  For the best neural model (RNNCRF), we report the analysis of feature importance using a similar approach to the one in \cite{Avati2017}. In short, we iterated over all features attached to the last
HF event, and computed the probability of readmission with a feature present or absent.
Computing the difference between both probabilities allowed us to quantify a feature's importance across the five folds. In the supplementary material section, we present additional variations
on this technique. Overall,
the average overlap (using Jaccard similarity) of the top-100 features between LASSO
and the RNNCRF model is 51\% and 55\% for increase and decrease of readmission probability, respectively.

\begin{table}
\begin{tabular}{|c|c|}
\hline 
Variables & HF Dataset (n=272,778)\tabularnewline
\hline 
\hline 
\textbf{Socio-demographics} & \tabularnewline
\hline 
Age, mean (SD) & 72.89 (14)\tabularnewline
\hline 
Gender female, count (\%) & 133765 (49\%)\tabularnewline
\hline 
\textbf{Pay source, count (\%)} & \tabularnewline
\hline 
Medicare & 391535 (76.4\%)\tabularnewline
\hline 
Private insurance & 47327 (9.23\%)\tabularnewline
\hline 
Medicaid & 47095 (9.19\%)\tabularnewline
\hline 
Self-pay & 13115 (2.55\%)\tabularnewline
\hline 
Other & 11859 (2.31\%)\tabularnewline
\hline 
No charge & 1514 (0.29\%)\tabularnewline
\hline 
\textbf{Hospitalization events} & \tabularnewline
\hline 
HF events, count (\%) & 343328 (66.94\%)\tabularnewline
\hline 
30 days all-cause readmission, count (\%) & 81087 (23.61\%)\tabularnewline
\hline 
Timeline length, mean (SD) & 1.88 (1.4)\tabularnewline
\hline 
\textbf{Top 5 diagnosis, count (\%)} & \tabularnewline
\hline 
Congestive heart failure; non-hypertensive & 777047 (10.29\%)\tabularnewline
\hline 
Coronary atherosclerosis and other heart disease & 547890 (7.25\%)\tabularnewline
\hline 
Residual codes & 305406 (4.04\%)\tabularnewline
\hline 
Cardiac dysrhythmias & 298823 (3.95\%)\tabularnewline
\hline 
Chronic kidney disease & 254593 (3.37\%)\tabularnewline
\hline 
\textbf{Top 5 procedures, count (\%)} & \tabularnewline
\hline 
Diagnostic cardiac catheterization; coronary arteriography & 106428 (14.95\%)\tabularnewline
\hline 
Respiratory intubation and mechanical ventilation & 57202 (8.03\%)\tabularnewline
\hline 
Blood transfusion & 52251 (7.34\%)\tabularnewline
\hline 
Diagnostic ultrasound of heart (echocardiogram) & 41076 (5.77\%)\tabularnewline
\hline 
Hemodialysis & 38083 (5.35\%)\tabularnewline
\hline 
\end{tabular}

\caption{\label{tab:Overview-of-HF}Overview of HF dataset}
\end{table}

\begin{table}
\begin{tabular}{|l|c|c|c|}
\hline 
Model name & AUC & CI - low & CI - high\tabularnewline
\hline 
\hline 
CNN & 0.619 & 0.616 & 0.622\tabularnewline
\hline 
CNN-Wide & 0.632 & 0.629 & 0.635\tabularnewline
\hline 
RNN (Convex\_HF\_lastHF) & 0.635 & 0.632 & 0.638\tabularnewline
\hline 
RNN (LastHF) & 0.636 & 0.633 & 0.638\tabularnewline
\hline 
RNN (Uniform\_HF) & 0.631 & 0.628 & 0.634\tabularnewline
\hline 
RNN (Convex\_HF\_NonHF) & 0.627 & 0.624 & 0.630\tabularnewline
\hline 
RNNSS (Convex\_HF\_lastHF) & 0.621 & 0.618 & 0.624\tabularnewline
\hline 
RNNSS (LastHF) & 0.625 & 0.623 & 0.628\tabularnewline
\hline 
RNNSS (Uniform\_HF) & 0.617 & 0.614 & 0.619\tabularnewline
\hline 
RNNSS (Convex\_HF\_NonHF) & 0.625 & 0.622 & 0.628\tabularnewline
\hline 
Neural CRF (Pairwise) & 0.634 & 0.631 & 0.637\tabularnewline
\hline 
Neural CRF (Unary) & 0.631 & 0.629 & 0.634\tabularnewline
\hline 
CRF Only (Pairwise) & 0.628 & 0.625 & 0.631\tabularnewline
\hline 
CRF Only (Unary) & 0.630 & 0.627 & 0.633\tabularnewline
\hline 
RNNCRF (Pairwise) & \textbf{0.642} & 0.640 & 0.645\tabularnewline
\hline 
RNNCRF (Unary) & 0.638 & 0.635 & 0.641\tabularnewline
\hline 
MLP & 0.628 & 0.625 & 0.631\tabularnewline
\hline 
Logistic regression (L2 reg.) & 0.637 & 0.634 & 0.640\tabularnewline
\hline 
Logistic regression (L1 reg.) & \textbf{0.643} & 0.640 & 0.646\tabularnewline
\hline 
\end{tabular}

\caption{\label{tab:AUC-models'-performance}AUC models' performance}
\end{table}

\section{Discussion}

This work highlights the advantages and limitations of deep learning
in the domain of HF readmission prediction. As a first result, we observe that deep learning profits from the incorporation
of patients' timeline / historical data for improving prediction performance.
Particularly, sequence labeling (RNN, RNNCRF, CRF only, Neural CRF)
and sequence classification (CNN-Wide) overall perform better than
event (non-timeline) classification (MLP). This finding is noteworthy,
as it reflects the ability of deep learning, with the help of fast
and highly specialized hardware and software, to utilize vast data
resources to produce increasingly state-of-the-art machine learning
performances. In the medical field, this amounts to a unique opportunity
to allow machines to base their predictions not only on the current
status of a patient, but also on the patient's history, and, if possible,
on the comparative analysis to all patient data (present and historical)
in a hospital system. Our data supports this notion, showing that
the detailed past history, reflected in a timeline of patient hospitalization
events, indeed carries additional information that boosts deep learning
prediction performance. Not all neural models are born equal though,
and for our study, we find that a scheduled sampling approach for
RNN did not improve timeline-based predictions. Interestingly, pairing
the RNN with a graphical model (CRF) resulted in the best performing
neural model, an observation that mirrors previous results in the
field of NLP\cite{Lample2016, Ma2016}. Similarly, neural CRF performed better than CRF alone, hinting at the importance of adding nonlinear features to graphical
models. While the actual performance numbers, with a maximal ROC score
of 0.642 AUC (95\% CI, 0.640-0.645), are inline with published machine-learning
predictions of HF readmission\cite{JD2017, Mortazavi2016}, it should be noted that they are based
on administrative data, rather than rich EHR data as used in earlier
studies. As such, our numbers represent the lower bound of achievable performance and deep learning on EHR data may eventually beat existing performance
numbers for readmission prediction, as it did in other areas such as diagnosis and disease prediction \cite{Choi2017,Lipton2015}. Nevertheless, the exact approach for attaining better performance remains
an open research question. One obvious avenue is to use multi-modal
learning, incorporating several clinical data sources (including images),
to offset the inherent issues with textual medical data, such as sparsity,
missingness, and incompleteness. Our second result addresses the question
of deep learning versus logistic regression for readmission prediction.
Our face-off shows that logistic regression with regularization matches
the best neural network performance. Our study attempted to compare
these two approaches as fairly as possible, allowing both methods
to perform hyperparameter optimization in the training phase, and
giving logistic regression, which uses data from the last hospitalization
event only, access to a patient's hospitalization history by adding
timeline summary data as an additional feature of the hospitalization
event. Nevertheless, the LASSO model had a couple of advantages over the neural models by (1) having access to the whole training set during the hyperparameter optimization, and (2) using $l_{1}$-norm regularization that served as feature-selection procedure while training the model. In contrast, the neural-based models had a very large set of hyperparameters to choose from (such as number of layers, dimensions of hidden vectors, etc.) that made it infeasible to explore the full hyperparameter space. We therefore opted for a random selection process for network configurations and optimization settings, and tested those against a subset of the training data only. Interestingly, and supporting our results, research by Rajkomar et al. \cite{Rajkomar2018} on a more general
hospital readmission problem (not focused on HF) also showed that logistic
regression with regularization (LASSO) is competitive compared
to an RNN model.  Focusing on neural models, we show that data contained in a patient's history boost prediction performance. Particularly, neural models showed higher performance with the length of a patient's timeline. EHRs contain historical information spanning several decades, and
we will test in future studies whether deep learning on timelines
greater than 11 month (the maximum used in this study) will succeed in besting the
performance of logistic regression model. Finally, our study sheds light on which features from the current or past hospitalizations are essential in readmission prediction. Features such as number of
hospitalization events, length of stay, thrombophlebitis/thromboembolism and
discharge against medical advice are among the highest contributing features to
the log-odds of readmission. Generally, diagnoses and administered procedures
pertaining to heart problems, such as contrast aortograms, result in increased
readmission probability, as does the number of comorbidities. Interestingly, particular payment sources (Medicare and Medicaid) are associated with increased, while self-pay is associated with decreased readmissions. \newline
In conclusion, we show that neural network models and logistic regression have comparable performance on HF readmission prediction using administrative data. We also demonstrate that the use of patient timeline data boosts the performance of neural models.

\paragraph*{Availability\newline}
The data processing, model implementation, training and testing workflow
in addition to the trained models are publicly available at https://bitbucket.org/A\_2/hcup\_research

\paragraph*{Acknowledgements\newline}
We would like to thank Dr. Harlan Krumholz for his insights and discussions about the research topic.

\paragraph*{Competing Interests\newline}
None.
\paragraph*{Funding\newline}
AA was supported by a fellowship from the Intramural  Research  Program of the National  Institutes  of  Health  (NIH), National Library of Medicine (NLM), and Lister Hill National Center for Biomedical Communications (LHNCBC).
\paragraph*{Contributions\newline}
AA and MK conceptualized and designed the experiments. MN and GT contributed in the design of the experiments. MK and GT supervised the project. AA worked on the development of processing and analysis workflow, algorithms and model implementation. MN contributed in the analysis workflow. AA and MK analyzed and interpreted the data. AA and MK wrote the manuscript.

\bibliographystyle{naturemag}
\bibliography{main_manuscript}
%%%%%%%%%% Merge with supplemental materials %%%%%%%%%%
\pagebreak
\begin{center}
\textbf{\large Supplementary Material for Neural networks versus Logistic
regression for 30 days all-cause readmission prediction}
\end{center}
\appendix
\section{Methods}
\subsection{Recurrent neural network (RNN)}

\begin{figure}
\includegraphics[scale=0.34]{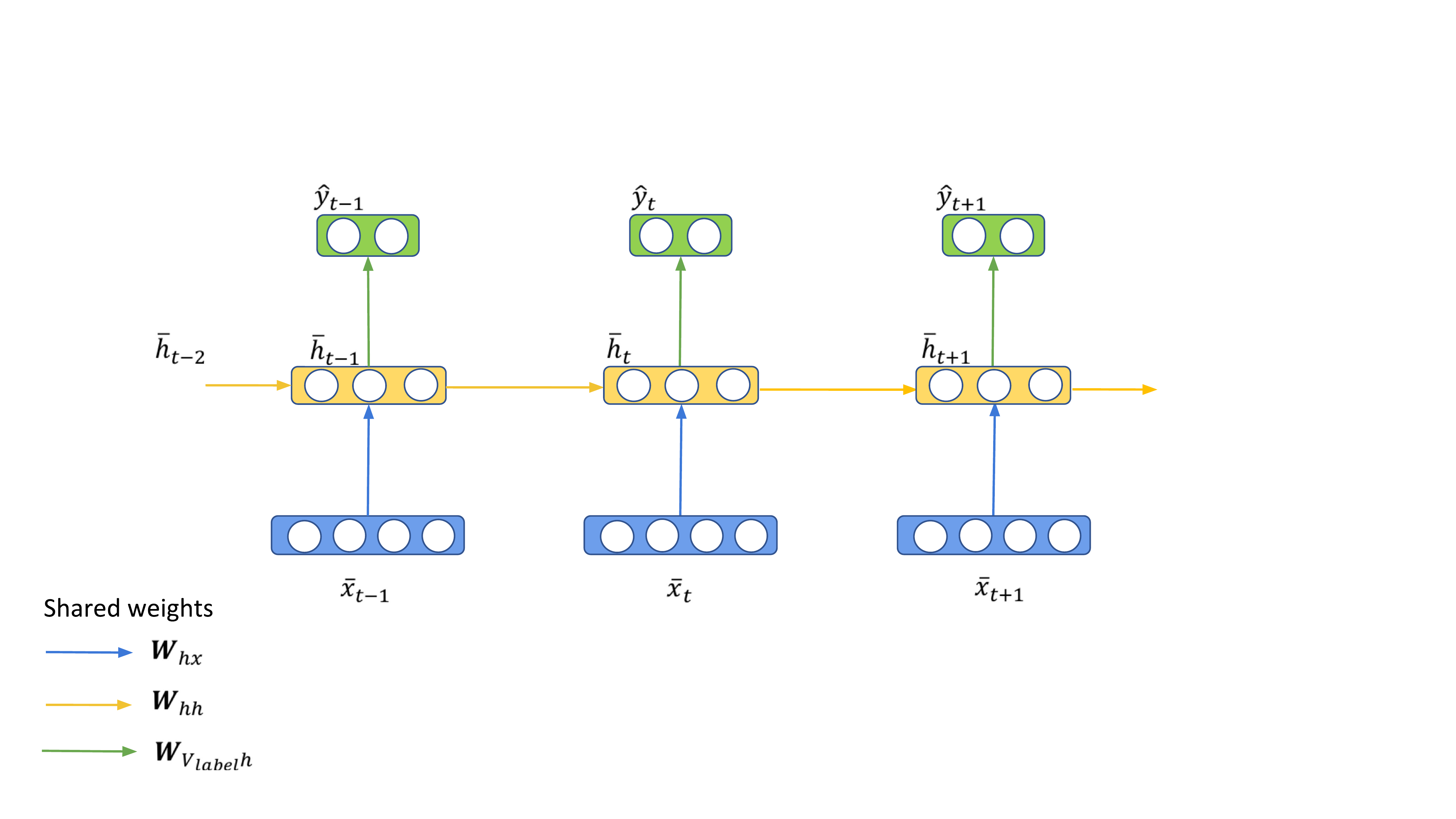}
\caption{\label{fig:Suppl_rnn-generic}Graphical representation of unfolded RNN}
\end{figure}

RNN computes a hidden vector at each time step (i.e. state vector
$\overline{h}_t$  at time $t$), representing a history or context
summary of the sequence using the input and hidden states vector
form the previous time step. This allows the model to learn long-range
dependencies where the network is unfolded as many times as the length
of the sequence it is modeling. Equation \ref{eq:Suppl_rnn_ht} shows the
computation of the hidden vector $\overline{h}_t$ using the input
$\overline{x}_t$ and the previous hidden vector $\overline{h}_{t-1}$
where $\phi$ is a non-linear transformation such as $ReLU(z)=max(0,z)$
or $tanh(z)=\frac{e^z - e^{-z}}{e^z + e^{-z}}$. To compute the outcome
$\hat{y}_t$ at time $t$, an affine transformation followed by non-linear
function are applied to the state vector $\overline{h}_t$ as
described in Equation \ref{eq:Suppl_rnn_yt}. The non-linear operator $\sigma$
can be either the $sigmoid$ function $\sigma(z) = \frac{1}{1+e^{-z}}$
applied to scalar input $z \in \R$, or its generalization the $softmax$ function applied to vector $\overline{z} \in \R^K$, $softmax(\overline{z})_i = \frac{e^{z_i}}{\sum_{j=1}^{K}e^{z_j}}$
for $i=1,...,K$. As a result, the outcome $\hat{y}_t$ represents
a probability distribution over the set of possible labels $V_{label}$
at time $t$. 

\begin{equation}
\overline{h}_{t}=\phi(\mathbf{W}_{hx}\overline{x}_{t}+\mathbf{W}_{hh}\overline{h}_{t-1}+\overline{b}_{hx})\label{eq:Suppl_rnn_ht}
\end{equation}

\begin{equation}
\hat{y}_{t}=\sigma(\mathbf{W}_{V_{label}h}\overline{h}_{t}+\overline{b}_{V_{label}})\label{eq:Suppl_rnn_yt}
\end{equation}
\newline
where $\mathbf W_{hh} \in \R^{D_h \times D_h}$, $\mathbf W_{hx} \in \R^{D_h \times d}$,
$\mathbf W_{V_{label}h} \in \R^{|V_{label}| \times D_h}$, $\overline{b}_{hx} \in \R^{D_h}$,
$\overline{b}_{V_{label}} \in \R^{|V_{label}|}$ representing the
model weights $\bm{\theta}$ to be optimized and $D_h$, $d$ are
the dimensions of $\overline{h}_t$ and $\overline{x}_t$ vectors
respectively. Note that the weights are shared across all the network (see Figure \ref{fig:Suppl_rnn-generic}
for RNN representation).

\subsubsection{Long short-term memory (LSTM)}

Long short-term memory (LSTM) \cite{Hochreiter1997,Graves2012} falls
in the gated memory cells approach that modifies the basic RNN by
replacing the standard neurons/units in the hidden layer with gated/memory cells to generate the hidden state vector $\overline{h}_t$
as described by the equations below. Moreover, LSTM introduces a new
cell state vector $\overline{c}_t$ that overall contributes in the
decision mechanism on what part of the history to keep or forget.
The computation of the output $\hat{y}_t$ at time $t$ remains the
same as explained in the RNN section.

\begin{alignat*}{2}
\overline{i}_{t} & =\sigma(\mathbf{W}_{hx}^{i}\overline{x}_{t}+\mathbf{W}_{hh}^{i}\overline{h}_{t-1}+\overline{b}_{hx}^{i}) & \qquad\qquad\qquad\text{(input gate)}\\
\overline{f}_{t} & =\sigma(\mathbf{W}_{hx}^{f}\overline{x}_{t}+\mathbf{W}_{hh}^{f}\overline{h}_{t-1}+\overline{b}_{hx}^{f}) & \qquad\qquad\qquad\text{(forget gate)}\\
\overline{o}_{t} & =\sigma(\mathbf{W}_{hx}^{o}\overline{x}_{t}+\mathbf{W}_{hh}^{o}\overline{h}_{t-1}+\overline{b}_{hx}^{o}) & \qquad\qquad\qquad\text{(output gate)}\\
\overline{\tilde{c}}_{t} & =\phi(\mathbf{W}_{hx}^{\tilde{c}}\overline{x}_{t}+\mathbf{W}_{hh}^{\tilde{c}}\overline{h}_{t-1}+\overline{b}_{hx}^{\tilde{c}}) & \qquad\qquad\text{\qquad(new state/memory cell)}\\
\overline{c}_{t} & =\overline{f}_{t}\odot\overline{c}_{t-1}+\overline{i}_{t}\odot\overline{\tilde{c}}_{t} & \qquad\qquad\qquad\text{(final cell state)}\\
\overline{h}_{t} & =\overline{o}_{t}\odot\phi(\overline{c}_{t}) & \qquad\qquad\qquad\text{(hidden state vector)}
\end{alignat*}
\newline
where $\mathbf{W}^i_{hx}$, $\mathbf{W}^f_{hx}$, $\mathbf{W}^o_{hx}$,
$\mathbf{W}^{\tilde{c}}_{hx}$ each $\in \R^{D_h \times d}$ and $\mathbf{W}^i_{hh}$,
$\mathbf{W}^f_{hh}$, $\mathbf{W}^o_{hh}$, $\mathbf{W}^{\tilde{c}}_{hh}$
each $\in \R^{D_h \times D_h}$. The biases $\overline{b}^{i}_{hx}$,
$\overline{b}^{f}_{hx}$, $\overline{b}^{o}_{hx}$, $\overline{b}^{\tilde{c}}_{hx}$
each $\in \R^{D_h}$ where $D_h$ and $d$ are the dimensions of $\overline{h}_t$
and $\overline{x}_t$ vectors respectively. The operator $\sigma$
represents the $sigmoid$ function, $\phi$ the $tanh$ or $ReLU$
function, and $\odot$ the element-wise product (i.e. Hadamard product)
function. Compared to the standard/conventional RNN (see Equation
\ref{eq:Suppl_rnn_ht}), it can be noted the added complexity in terms of
the number of weight matrices and biases required to compute the hidden
state vector $\overline{h}_t$.

\subsubsection{Gated recurrent unit (GRU)}

Gated recurrent unit (GRU) \cite{Cho2014} presents similar approach
to LSTM but with a simpler model that modifies the computation mechanism
of the hidden state vector $\overline{h}_t$ through the specified
equations below.

\begin{alignat*}{2}
\overline{z}_{t} & =\sigma(\mathbf{W}_{hx}^{z}\overline{x}_{t}+\mathbf{W}_{hh}^{z}\overline{h}_{t-1}+\overline{b}_{hx}^{z}) & \qquad\qquad\qquad\text{(update gate)}\\
\overline{r}_{t} & =\sigma(\mathbf{W}_{hx}^{r}\overline{x}_{t}+\mathbf{W}_{hh}^{r}\overline{h}_{t-1}+\overline{b}_{hx}^{r}) & \qquad\qquad\qquad\text{(reset gate)}\\
\overline{\tilde{h}}_{t} & =\phi(\mathbf{W}_{hx}^{\tilde{h}}\overline{x}_{t}+\overline{r}_{t}\odot\mathbf{W}_{hh}^{\tilde{h}}\overline{h}_{t-1}+\overline{b}_{hx}^{\tilde{h}}) & \qquad\qquad\text{\qquad(new state/memory cell)}\\
\overline{h}_{t} & =(1-\overline{z}_{t})\odot\overline{\tilde{h}}_{t}+z_{t}\odot h_{t-1} & \qquad\qquad\qquad\text{(hidden state vector)}
\end{alignat*}\newline
The model computes a reset gate $\overline{r}_t$ that is used to
modulate the effect of the previous hidden state vector $\overline{h}_{t-1}$
when computing the new memory vector $\overline{\tilde{h}}_t$. The
update gate $\overline{z}_t$ determines the importance/contribution
of the newly generated memory vector $\overline{\tilde{h}}_t$ compared
to the previous hidden state vector $\overline{h}_{t-1}$ when computing
the current hidden vector $\overline{h}_t$. The weights $\mathbf{W}^z_{hx}$,
$\mathbf{W}^r_{hx}$, $\mathbf{W}^{\tilde{h}}_{hx}$ each $\in \R^{D_h \times d}$
and $\mathbf{W}^z_{hh}$, $\mathbf{W}^r_{hh}$, $\mathbf{W}^{\tilde{h}}_{hh}$
each $\in \R^{D_h \times D_h}$. The biases $\overline{b}^{z}_{hx}$,
$\overline{b}^{r}_{hx}$, $\overline{b}^{\tilde{h}}_{hx}$ each $\in \R^{D_h}$
where $D_h$ and $d$ are the dimensions of $\overline{h}_t$ and
$\overline{x}_t$ vectors respectively. The operators notation have
the same meaning as described in the LSTM section. 

\subsubsection{RNN objective function\label{subsec:Suppl_RNN-objective-function}}

In our work, we use RNN to refer to RNN, LSTM and GRU models targeting
the sequence labeling view of the problem. We defined the loss for
an $i$-th sequence at each time step by the cross-entropy error

\begin{equation}
l_{t}^{(i)}=-\sum_{c=1}^{|V_{label}|}y_{t,c}^{(i)}\times log(\hat{y}_{t,c}^{(i)})
\end{equation}
where the loss for the $i$-th sequence (i.e. $i$-th patient's timeline/trajectory)
is defined by the average loss over the sequence length $T_i$

\begin{equation}
L_{i}=\frac{1}{T_{i}}\sum_{t=1}^{T_{i}}l_{t}^{(i)}\label{eq:Suppl_loss_seq}
\end{equation}
Given that our focus is on the 30 days all-cause readmissions after
HF hospitalization, our model's focus should be on the index events
(i.e. claims/events where HF is the primary diagnosis of hospitalization).
Hence, the objective function could be modified to reflect this requirement.
We modify the defined loss over $i$-th sequence (see Equation \ref{eq:Suppl_loss_seq})
by defining an average loss for non-index and index events separately
and then taking a convex combination between both losses parametrized
by $\alpha$. The parameter $\alpha$ is determined using a validation
set. Our modification is inspired by the work done in \cite{Lipton2015}.

\[
L_{i}^{HF}=\frac{1}{\sum_{t=1}^{T_{i}}\mathbbm{1}[\overline{x}_{t,primaryHF}^{(i)}=1]}\sum_{t=1}^{T_{i}}l_{t}^{(i)}\mathbbm{1}[\overline{x}_{t,primaryHF}^{(i)}=1]
\]

\[
L_{i}^{nonHF}=\frac{1}{\sum_{t=1}^{T_{i}}\mathbbm{1}[\overline{x}_{t,primaryHF}^{(i)}=0]}\sum_{t=1}^{T_{i}}l_{t}^{(i)}\mathbbm{1}[\overline{x}_{t,primaryHF}^{(i)}=0]
\]

\begin{equation}
L_{i}=(1-\alpha)L_{i}^{nonHF}+\alpha(L_{i}^{HF})\label{eq:Suppl_rnn_objfunc}
\end{equation}
\newline
where $\mathbbm{1}[\overline{x}_{t,primaryHF}^{(i)}=1]$ is an indicator
function that is equal to $1$ when the feature vector $\overline{x}_{t}^{(i)}$
representing the event at time $t$ for the $i$-th sequence
has HF as the primary diagnosis (we refer to this loss by Convex\_HF\_NonHF). A second variation (Uniform\_HF) to the first objective
function in Equation \ref{eq:Suppl_rnn_objfunc}, is to consider only the index
events in the patients' timeline where we compute the average
cross-entropy loss for the index events only.

\begin{equation}
L_{i}=L_{i}^{HF}\label{eq:Suppl_rnn_objfunc_2}
\end{equation}
A third variation (Convex\_HF\_lastHF) is to take a convex
combination between all index events and the last index event in the
timeline. 

\[
L_{i}^{HF}=\frac{1}{\sum_{t=1}^{T_{i}}\mathbbm{1}[\overline{x}_{t,primaryHF}^{(i)}=1]}\sum_{t=1}^{T_{i}}l_{t}^{(i)}\mathbbm{1}[\overline{x}_{t,primaryHF}^{(i)}=1]
\]

\[
L_{i}^{lastHF}=l_{T}^{(i)}
\]

\begin{equation}
L_{i}=(1-\alpha)L_{i}^{HF}+\alpha(L_{i}^{lastHF})\label{eq:Suppl_rnn_objfunc_3}
\end{equation}
\newline
where $L_{i}^{lastHF}$ is the cross-entropy loss for the last HF
event (i.e. the last index event in the patient's timeline) and $l_{T}^{(i)}=-\sum_{c=1}^{|V_{label}|}y_{T,c}^{(i)}\times log(\hat{y}_{T,c}^{(i)})$.
\newline The previous variations consider the \textit{sequence labeling}
approach since the loss function for a sequence is defined as a composite
of losses from different times/events in a patient's timeline. A final
variation that uses the \textit{sequence classification} view of the
problem is to define an objective function that focuses only on the
last HF event (LastHF) by computing the loss at the last target event we aim to predict.

\begin{equation}
L_{i}=L_{i}^{lastHF}\label{eq:Suppl_rnn_objfunc_4}
\end{equation}
Lastly, the objective function for the whole training set $D_{train}$
is defined by the average loss across all the sequences in $D_{train}$
plus a weight regularization term (i.e. $l_2$-norm regularization)
applied to the model parameters represented by $\bm{\theta}$

\begin{equation}
L(\mathbf{\bm{\theta}})=\frac{1}{N}\sum_{i=1}^{N}L_{i}+\frac{\lambda}{2}||\mathbb{\bm{\theta}}||_{2}^{2}\label{eq:Suppl_trainnn_obj}
\end{equation}
In practice, the training occurs using mini-batches where computing
the loss function and updating the parameters/weight occur after processing
each mini-batch of the training set.

\subsubsection{RNN with conditional random fields (CRF)}

CRF models the conditional probability of a sequence $\mathbf{\underline{y}}$
given its corresponding sequence of observation vectors $\mathbf{\underline{x}}$
(i.e. $p(y_{1},y_{2},...,y_{T}|\overline{x}_{1},\overline{x}_{2},...,\overline{x}_{T})$)
using a parametrized \textit{global} feature vector $\overline{F}(\mathbf{\underline{x},}\mathbf{\underline{y})}\in\R^{J}$
that takes input/output sequences to produce $J$-dimensional vector.
As a result, the computation of the conditional probability of an
output sequence given its input sequence of observations is equal
to 

\begin{equation}
p(\mathbf{\underline{y}}|\mathbf{\underline{x}})=\frac{e^{\bm{\theta}\cdot\overline{F}(\mathbf{\underline{x},\mathbf{\underline{y})}}}}{\sum_{\mathbf{\underline{y}^{'}\in\mathbf{\underline{Y}}}}e^{\bm{\theta}\cdot\overline{F}(\mathbf{\underline{x},\mathbf{\underline{y}')}}}}
\end{equation}
where $\mathbf{\underline{Y}}$ is the set of all label sequences,
the denominator represents the normalizer (commonly referred to the
partition function $Z$), and $\bm{\theta}$ is the weight vector
corresponding to the global feature vector $\overline{F}(\mathbf{\underline{x},}\mathbf{\underline{y})}$.
The common definition of the feature vector $\overline{F}$ uses the
first-order Markov assumption in order to make the inference and model
training tractable \cite{Collins}. That is 

\begin{equation}
\overline{F}(\mathbf{\underline{x},}\mathbf{\underline{y})}=\sum_{t=1}^{T}\overline{f}(\mathbf{\underline{x}},t,y_{t-1},y_{t})
\end{equation}
where the global feature vector $\overline{F}$ is the sum of a local
feature vector $\overline{f}$ applied at each time step until the
end of the sequence. The local vector $\overline{f}$ has the same
dimension of $\overline{F}$ (i.e. $\in\R^{J}$) and has access to
the whole observation sequence $\mathbf{\underline{x}}$, and the
current and previous states/outputs $y_{t-1}$ and $y_{t}$ \cite{Collins}.
Generally, increasing the model order $k$ (i.e. $k\ge2$) would lead
to exponential computational complexity in terms of $k$. However,
recent work as in \cite{Ye2009,Cuong2014,Vieira2016}, showed under
the assumption of label pattern sparsity, the use of higher-order
models (i.e. models with $k\ge2$) is feasible without incurring an
exponential complexity in the training and inference algorithms \cite{Allam2017}.
\newline We denote the output features of the RNN layer by $\mathbf{\underline{z}}=[\overline{z}_{1},\overline{z}_{2},\cdots,\overline{z}_{T}]$
representing the sequence of output features computed from the input
sequence $\mathbf{\underline{x}}$ (both sequences have equal length).
The potential functions in the CRF layer are computed using $\mathbf{\underline{z}}$
along with label sequence $\mathbf{\underline{y}}$. In our work,
we experimented with two potential functions: 
\begin{enumerate}
\item RNNCRF (Unary) that computes unary potentials $\psi_{y_{t}}(\overline{z}_{t})$
by passing the RNN output feature vector $\overline{z}_{t}$ at time
$t$ to a linear affine map and applying a non-linear transformation
resulting in a vector of size equal to the number of classes $|V_{label}|$
for each $\overline{z}_{t}$. The pairwise potential is modeled using
a transition parameters matrix $A(y_{t-1},y_{t})$ of size $|V_{label}|\times|V_{label}|$
representing the transition score from one outcome class to another.
The total score computation is equal to $\overline{F}(\mathbf{\underline{z},}\mathbf{\underline{y})}=\sum_{t=1}^{T}(\psi_{y_{t}}(\overline{z}_{t})+A(y_{t-1},y_{t})$). 
\item RNNCRF (Pairwise) that computes pairwise potentials $\psi_{y_{t-1}y_{t}}(\overline{z}_{t})$
by using linear affine map transformation followed by non-linear element-wise
operation generating an output of size $|V_{label}|\times|V_{label}|$
similar to the approach reported in \cite{Ma2016}. The total score $\overline{F}(\mathbf{\underline{z}},\mathbf{\underline{y}})$
is equal to $\overline{F}(\mathbf{\underline{z},}\mathbf{\underline{y})}=\sum_{t=1}^{T}\psi_{y_{t-1}y_{t}}(\overline{z}_{t})$.
\newline
\end{enumerate}

\subsection{Dataset features representation\label{subsec:Suppl_Dataset-features-representation}}

\subsubsection{Input features $\overline{x}_{t}$}

Each claim/event in a patient's timeline is represented by a feature
vector $\overline{x}_{t}$ encoding the characteristics of the hospitalization
event and the corresponding patient. The feature vector $\overline{x}_{t}$
is composed of:
\begin{itemize}
\item Diagnosis: every claim in the dataset includes 25 ordered fields,
each registering patient's diagnosis category based on CCS grouper
\cite{AgencyforHealthcareResearchandQuality2009} during the corresponding
hospitalization event. We first extracted set $V_{diagnosis}$ representing
the diagnosis having at least 1000 counts/occurrences registered in the HF dataset. Then, we constructed the following vectors: 
\begin{enumerate}
\item $\overline{x}_{diag1}$ a one-hot encoded vector $\in\{0,1\}^{|V_{diagnosis}|}$
representing the diagnosis category registered for the primary diagnosis
field 
\item $\overline{x}_{diag2}$ a one-hot encoded vector $\in\{0,1\}^{|V_{diagnosis}|}$
representing the diagnosis category registered for the secondary diagnosis
field
\item $\overline{x}_{diag3}$ a one-hot encoded vector $\in\{0,1\}^{|V_{diagnosis}|}$
representing the diagnosis category registered for the tertiary diagnosis
field
\item $\overline{x}_{countdiag}$ a vector $\in\R^{|V_{diagnosis}|}$ representing
the count of diagnosis categories registered in all 25 diagnosis fields
\end{enumerate}
\item Procedures: every claim in the dataset includes 15 ordered fields,
each registering patient's administered procedure category based on
CCS grouper during the corresponding hospitalization event. We first
extracted set $V_{procedures}$ representing the top procedures having
at least 1000 counts registered in HF dataset. Then, we constructed
the following vectors: 
\begin{enumerate}
\item $\overline{x}_{proc1}$ a one-hot encoded vector $\in\{0,1\}^{|V_{procedures}|}$
representing the procedure category registered for the primary procedure
field
\item $\overline{x}_{proc2}$ a one-hot encoded vector $\in\{0,1\}^{|V_{procedures}|}$
representing the procedure category registered for the secondary procedure
field
\item $\overline{x}_{proc3}$ a one-hot encoded vector $\in\{0,1\}^{|V_{procedures}|}$
representing the procedure category registered for the tertiary procedure
field
\item $\overline{x}_{countproc}$ a vector $\in\R^{|V_{procedures}|}$ representing
the count of procedure categories registered in all 15 procedure fields
\end{enumerate}
\item Body-system chronic condition: every claim in the dataset includes
25 ordered fields, each representing body-system chronic condition
indicators, categorizing ICD-9-CM diagnosis codes into chronic or
not \cite{AgencyforHealthcareResearchandQuality2015}. We refer to
the list of body-system categories by set $V_{bchronic}$ that includes
18 categories. We constructed the following vectors: 
\begin{enumerate}
\item $\overline{x}_{bchronic1}$ a one-hot encoded vector $\in\{0,1\}^{|V_{bchronic}|}$
representing the body-system chronic condition indicator category
listed in the primary field
\item $\overline{x}_{bchronic2}$ a one-hot encoded vector $\in\{0,1\}^{|V_{bchronic}|}$
representing the body-system chronic condition indicator category
listed in the secondary field
\item $\overline{x}_{bchronic3}$ a one-hot encoded vector $\in\{0,1\}^{|V_{bchronic}|}$
representing the body-system chronic condition indicator category
listed in the tertiary field
\item $\overline{x}_{countbchronic}$ a vector $\in\R^{|V_{bchronic}|}$
representing the count of body-system chronic condition indicator
categories registered in the 25 fields
\end{enumerate}
\item External cause of injury code: every claim in the dataset includes
4 ordered fields, each detailing an injury code (E-code) based on
CCS software categorizing all ICD-9-CM diagnosis codes into 20 categories
\cite{AgencyforHealthcareResearchandQuality2015}. We refer to the
list of E-code categories by set $V_{ecode}$ that includes 20 categories.
We constructed the following vectors: 
\begin{enumerate}
\item $\overline{x}_{ecode1}$ a one-hot encoded vector $\in\{0,1\}^{|V_{ecode}|}$
representing the E-code injury category listed in the primary field
\item $\overline{x}_{countecode}$ a vector $\in\R^{|V_{ecode}|}$ representing
the count of E-code injury categories registered in the 4 fields
\end{enumerate}
\item Procedure classes: every claim in the dataset includes 15 ordered
fields, each describing a broad category code (i.e. class) based on
categorization of the ICD-9-CM procedure codes\cite{AgencyforHealthcareResearchandQuality2015}.
We refer to the list of procedure broad categories by set $V_{procedureclass}$
that includes 4 categories. We constructed the following vectors: 
\begin{enumerate}
\item $\overline{x}_{countpclass}$ a vector $\in\R^{|V_{procedureclass}|}$
representing the count of procedure class categories registered in
the 15 fields
\end{enumerate}
\item Comorbidity condition: every claim in the dataset includes 29 binary
fields $\in\{0,1\}$, each representing an indicator of a specific
comorbidity that was determined by the AHRQ comorbidity software \cite{HCUPSoftware2008}.
The software determines comorbidities that are more likely present
prior to hospitalization event \cite{HCUPSoftware2008}. We refer
to the comorbidity categories encoded in the 29 binary variables by
set $V_{comorbid}$. We constructed the following vector: 
\begin{enumerate}
\item $\overline{x}_{comorbid}$ a vector $\in\{0,1\}^{|V_{comorbid}|}$
where each component corresponds to one of the 29 binary fields representing
the presence/absence of a comorbidity condition
\end{enumerate}
\item Major diagnostic category (MDC) assigned by HCFA DRG Grouper algorithm
during the processing of HCUP dataset \cite{AgencyforHealthcareResearchandQuality2015}.
We refer to the list of MDC categories by set $V_{mdc}$ where we
constructed the following vector:
\begin{enumerate}
\item $\overline{x}_{mdc}$ a one-hot encoded vector $\in\{0,1\}^{|V_{mdc}|}$
representing the MDC code/category
\end{enumerate}
\item Risk of mortality subclass: every claim in the dataset includes a
field that measures risk of mortality subclass based on all patient
refined diagnosis related groups assigned using software developed
by 3M Health Information System \cite{AgencyforHealthcareResearchandQuality2015}.
The measure consists of five categories which we refer to by the set
$V_{riskmortal}$ . We constructed the following vector: 
\begin{enumerate}
\item $\overline{x}_{riskmortal}$ a one-hot encoded vector $\in\{0,1\}^{|V_{riskmortal}|}$
representing the risk of mortality subclass category
\end{enumerate}
\item Severity of illness subclass: every claim in the dataset includes
a field that measures severity of illness subclass based on all patient
refined diagnosis related groups assigned using software developed
by 3M Health Information System \cite{AgencyforHealthcareResearchandQuality2015}.
The measure consists of five categories which we refer to set $V_{severity}$
. We constructed the following vector: 
\begin{enumerate}
\item $\overline{x}_{severity}$ a one-hot encoded vector $\in\{0,1\}^{|V_{severity}|}$
representing the severity of illness subclass category
\end{enumerate}
\item Major operating room procedure indicator: $x_{orproc}\in\{0,1\}$
a binary variable indicating whether a major operating room procedure
was reported on discharge
\item Number of chronic conditions: $x_{nchronic}\in\R$ a variable representing
the counts of unique chronic diagnosis reported on the discharge
\item Socio-demographics: every claim is associated with a patient and includes
information regarding patient's age, gender, income and place/location
of residence. We constructed a vector $\overline{x}_{sociodem}$ that
represents the concatenation of the following variables:
\begin{enumerate}
\item age: $x_{age}\in\R$ a variable representing the age of the patient
\item gender: $x_{gender}\in\{0,1\}$ a binary variable representing the
gender of a patient
\item income: $\overline{x}_{income}\in\{0,1\}^{|V_{income}|}$ a one-hot
encoded vector indicating the median household income quartile for
patient's zip code, where $V_{income}$ is the set of income categories
\item place/location: $\overline{x}_{ploc}\in\{0,1\}^{|V_{ploc}|}$ a one-hot
encoded vector describing patient's location based on the National
Center for Health Statistics (NCHS) classification scheme for US counties,
where $V_{ploc}$ is the set of location categories
\item resident: $x_{resident}\in\{0,1\}$ a binary variable representing
if the patient is resident in the state in which they were treated
\end{enumerate}
\item Event/claim info: every claim included the length of stay of the hospitalization
event, if the admission was on a weekend, and the discharge month.
We also computed the time difference between the hospital admission
of a current claim/event and the discharge of previous claim/event
for all events in a patient's timeline. We constructed a vector $\overline{x}_{event}$
that is the concatenation of the following variables:
\begin{enumerate}
\item length of stay: $x_{los}\in\R$ a variable representing the length
of stay in days for each hospitalization event
\item time difference between consecutive events: $x_{\Delta t}\in\R$ a
variable representing the time difference in days between current
admission and previous discharge events
\item admission on weekend: $x_{aweekend}\in\{0,1\}$ a binary variable
indicating if a patient was admitted on a weekend
\item discharge month: $\overline{x}_{dmonth}\in\{0,1\}^{|V_{dmonth}|}$
a one-hot encoded vector indicating a patient's discharge month where
$V_{dmonth}$ is the set of recorded months in the dataset
\item disposition of patient: $\overline{x}_{dispuniform}\in\{0,1\}^{|V_{dispuniform}|}$
a one-hot encoded vector indicating the disposition of the patient
at discharge
\item expected primary payer: $\overline{x}_{paysrc}\in\{0,1\}^{|V_{paysrc}|}$
a one-hot encoded vector indicating the expected primary payer (such
as Medicare, private insurance, etc.)
\item same-day event: $\overline{x}_{sameday}\in\{0,1\}^{|V_{sameday}|}$
a one-hot encoded vector identifying transfers and/or same-day stay
collapsed records
\item elective admission: $x_{elective}\in\{0,1\}$ a binary variable indicating
elective versus non-elective admission
\item rehab transfer: $x_{rehab}\in\{0,1\}$ a binary variable indicating
if the claim is a combined record involving transfer to rehabilitation,
evaluation, or other aftercare
\item number of index events: $x_{countindex}\in\R$ a variable representing
the number of index events in the timeline of a patient up to the
current admission event (inclusive)
\item number of admission events: $x_{countevents}\in\R$ a variable representing
the number of admission events in the timeline of a patient up to
the current admission event (inclusive)
\end{enumerate}
\end{itemize}
Hence, the feature vector $\overline{x}_{t}$ is the concatenation
of all these variables, encoding the characteristics of both the event
and its corresponding patient at one time step of the patient's trajectory. 

\section{Experiments}

\subsection{Hyperparameters optimization}

\subsubsection{RNN model}

\begin{figure}
\includegraphics[scale=0.06]{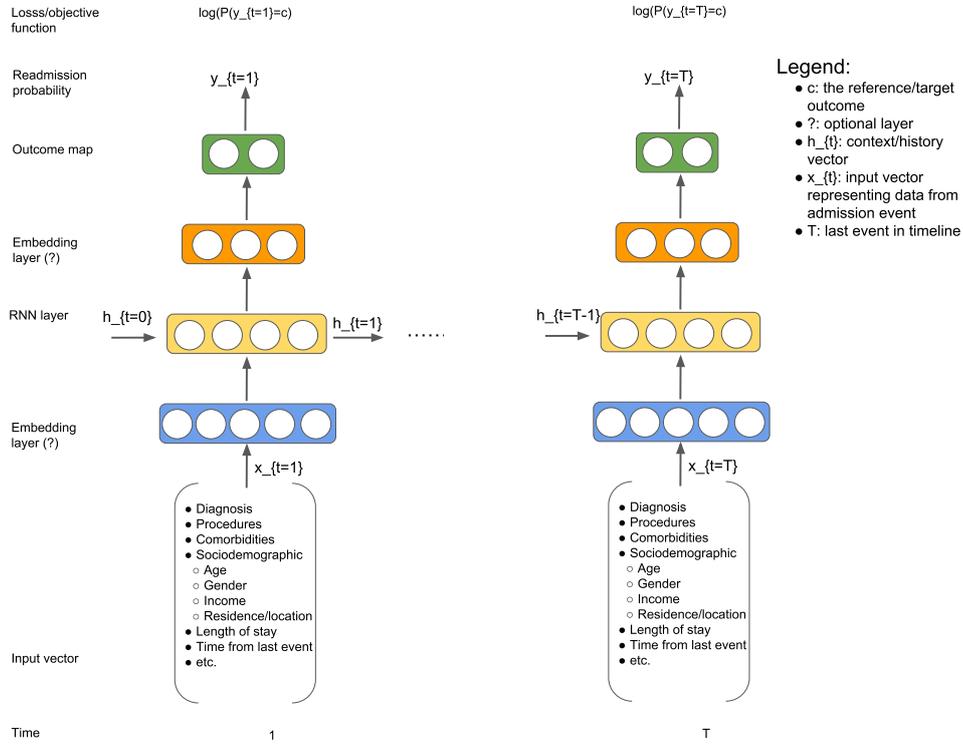}

\caption{\label{fig:Suppl_RNN-generic-model/architecture}RNN generic model/architecture}

\end{figure}

The set of all possible hyperparameters configuration (i.e. choice
of values for hyperparameters) for RNN models is reported in Table
\ref{tab:Suppl_RNN-hyperparameter-options}. These hyperparameters controlled
the network architecture design that is represented in Figure \ref{fig:Suppl_RNN-generic-model/architecture}.

\begin{table}
\resizebox{\textwidth}{!}{

\begin{tabular}{|c|c|c|}
\hline 
Parameter name  & Set/range values & Best/optimal value\tabularnewline
\hline 
\hline 
Embedding layer (Blue block) dimension & $\{0,\lfloor d/2\rfloor,\lfloor d/3\rfloor,\lfloor d/4\rfloor\}$
where $d$ is input dimension & $0$\tabularnewline
\hline 
RNN layer (Yellow block) &  & \tabularnewline
\hline 
RNN type & $\{\text{LSTM},\text{GRU},\text{Vanilla RNN}\}$ & Vanilla RNN \tabularnewline
\hline 
Hidden vector dimension $D_{h}$ & $\{8,16,32,64,128,256\}$ & $16$ \tabularnewline
\hline 
Number of hidden layers & $\{1,2,3\}$ & $1$ \tabularnewline
\hline 
Dropout probability $p_{dropout}$ & \{$0.15,0.35,0.5\}$ & $0.35$ \tabularnewline
\hline 
Embedding layer (Orange block) & $\{0,D_{h},\lfloor D_{h}/2\rfloor,\lfloor D_{h}/3\rfloor,\lfloor D_{h}/4\rfloor\}$ & $0$ \tabularnewline
\hline 
Non-linear function & $\{tanh,ReLU\}$ & $ReLU$ \tabularnewline
\hline 
$l_2$-norm regularization $\lambda$ & $\{10^{-3},10^{-2},10^{-1}\}$ & $10^{-2}$ \tabularnewline
\hline 
Convex combination parameter for the RNN objective function, $\alpha$ & $\{0.65,0.8,0.95\}$ & $0.8$ \tabularnewline
\hline 
Batch size during training $|B|$ & $\{8,16,32,64,128\}$ & $64$ \tabularnewline
\hline 
Optimization algorithm & \{Adam\} & Adam \tabularnewline
\hline 
\end{tabular}

}

\caption{\label{tab:Suppl_RNN-hyperparameter-options}RNN hyperparameter options
(see Figure \ref{fig:Suppl_RNN-generic-model/architecture})}
\end{table}

\subsubsection{RNNSS model}

The hyperparameters configurations for RNNSS model is reported in
Table \ref{tab:Suppl_RNNSS-hyperparameter-options}, which controlled the network architecture design depicted in Figure \ref{fig:Suppl_RNN-generic-model/architecture}.

\begin{table}
\resizebox{\textwidth}{!}{

\begin{tabular}{|c|c|c|}
\hline 
Parameter name  & Set/range values & Best/optimal value\tabularnewline
\hline 
\hline 
Embedding layer (Blue block) dimension & $\{0,\lfloor d/2\rfloor,\lfloor d/3\rfloor,\lfloor d/4\rfloor\}$
where $d$ is input dimension & $\lfloor d/2\rfloor$ \tabularnewline
\hline 
RNN layer (Yellow block) &  & \tabularnewline
\hline 
RNN type & $\{\text{LSTM},\text{GRU},\text{Vanilla RNN}\}$ & GRU \tabularnewline
\hline 
Hidden vector dimension $D_{h}$ & $\{8,16,32,64,128,256\}$ & $128$ \tabularnewline
\hline 
Number of hidden layers & $\{1,2,3\}$ & $1$ \tabularnewline
\hline 
Dropout probability $p_{dropout}$ & \{$0.15,0.35,0.5\}$ & $0.15$ \tabularnewline
\hline 
Embedding layer (Orange block) & $\{0,D_{h},\lfloor D_{h}/2\rfloor,\lfloor D_{h}/3\rfloor,\lfloor D_{h}/4\rfloor\}$ & $\lfloor D_{h}/3\rfloor$ \tabularnewline
\hline 
Non-linear function & $\{tanh,ReLU\}$ & $tanh$ \tabularnewline
\hline 
$l_2$-norm regularization $\lambda$ & $\{10^{-3},10^{-2},10^{-1}\}$ & $10^{-2}$ \tabularnewline
\hline 
Convex combination parameter for the RNN objective function, $\alpha$ & $\{0.65,0.8,0.95\}$ & $0.8$ \tabularnewline
\hline 
Batch size during training $|B|$ & $\{8,16,32,64,128\}$ & $64$ \tabularnewline
\hline 
Scheduled sampling parameter $\rho$ & \{Linear, Exponential, Sigmoid\} & Exponential \tabularnewline
\hline 
Optimization algorithm & \{Adam\} & Adam \tabularnewline
\hline 
\end{tabular}

}

\caption{\label{tab:Suppl_RNNSS-hyperparameter-options}RNNSS hyperparameter options
(see Figure \ref{fig:Suppl_RNN-generic-model/architecture})}
\end{table}

\subsubsection{RNNCRF model}

\begin{figure}
\includegraphics[scale=0.06]{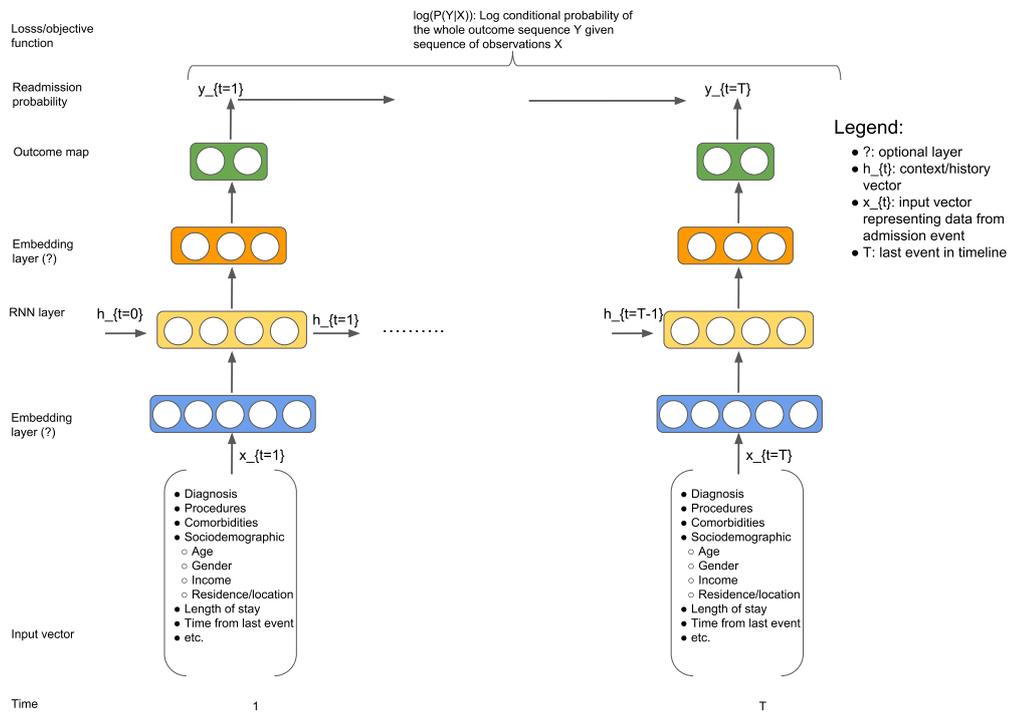}

\caption{\label{fig:Suppl_RNNCRF-generic-model/architecture}RNNCRF generic model/architecture}
\end{figure}

Similar to RNN models, the set of all possible hyperparameters configuration
for models using RNN with CRF is reported in Table \ref{tab:Suppl_RNNCRF-hyperparameter-options}
along with the network architecture design in Figure \ref{fig:Suppl_RNNCRF-generic-model/architecture}.

\begin{table}
\resizebox{\textwidth}{!}{

\begin{tabular}{|c|c|c|}
\hline 
Parameter name & Set/range values & Best/optimal value\tabularnewline
\hline 
\hline 
Embedding layer (Blue block) dimension & $\{0,\lfloor d/2\rfloor,\lfloor d/3\rfloor,\lfloor d/4\rfloor\}$
where $d$ is input dimension & $\lfloor d/2\rfloor$ \tabularnewline
\hline 
RNN layer (Yellow block) &  & \tabularnewline
\hline 
RNN type & $\{\text{LSTM},\text{GRU},\text{Vanilla RNN}\}$ & GRU \tabularnewline
\hline 
Hidden vector dimension $D_{h}$ & $\{8,16,32,64,128,256\}$ & $128$ \tabularnewline
\hline 
Number of hidden layers & $\{1,2,3\}$ & $1$ \tabularnewline
\hline 
Dropout probability $p_{dropout}$ & \{$0.15,0.35,0.5\}$ & $0.15$ \tabularnewline
\hline 
Embedding layer (Orange block) & $\{0,D_{h},\lfloor D_{h}/2\rfloor,\lfloor D_{h}/3\rfloor,\lfloor D_{h}/4\rfloor\}$ & $\lfloor D_{h}/3\rfloor$ \tabularnewline
\hline 
Non-linear function & $\{tanh,ReLU\}$ & $tanh$ \tabularnewline
\hline 
$l_2$-norm regularization $\lambda$ & $\{10^{-3},10^{-2},10^{-1}\}$ & $10^{-2}$ \tabularnewline
\hline 
Batch size during training $|B|$ & $\{8,16,32,64,128\}$ & $64$ \tabularnewline
\hline 
Optimization algorithm & \{Adam\} & Adam \tabularnewline
\hline 
\end{tabular}

}

\caption{\label{tab:Suppl_RNNCRF-hyperparameter-options}RNNCRF hyperparameter options
(see Figure \ref{fig:Suppl_RNNCRF-generic-model/architecture})}
\end{table}

\subsubsection{CRF and Neural CRF models}

\begin{figure}
\includegraphics[scale=0.06]{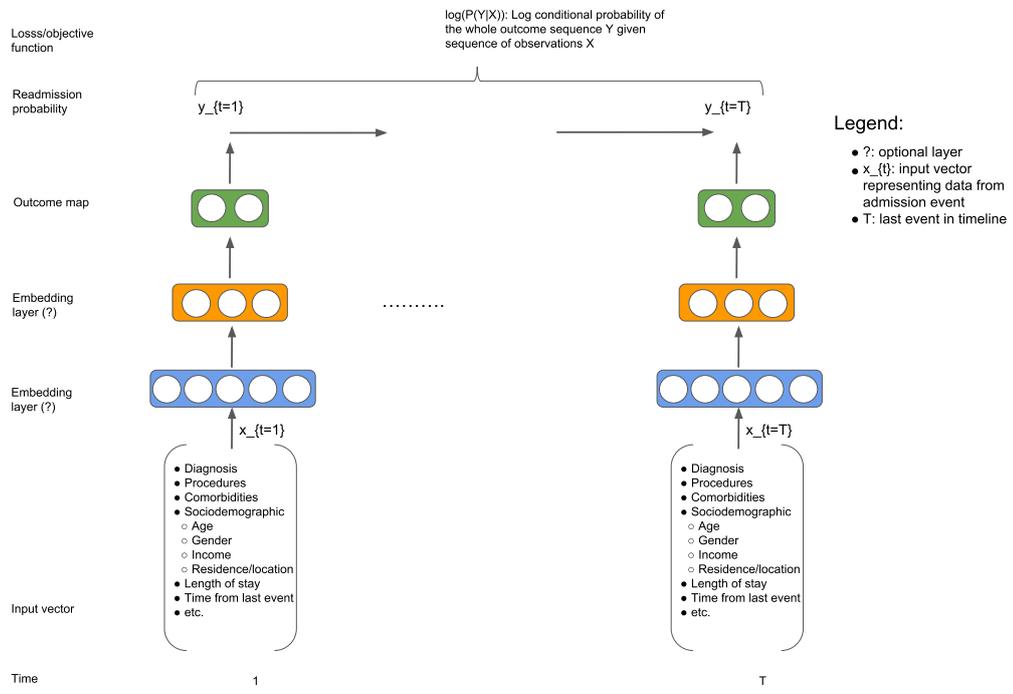}

\caption{\label{fig:Suppl_CRF-generic-model/architecture}CRF generic model/architecture}
\end{figure}

CRF only and Neural CRF models' hyperparameters configuration is reported
in Tables \ref{tab:Suppl_CRF-hyperparameter-options} and \ref{tab:Suppl_CRFNeural-hyperparameter-options} respectively along with the network architecture/design in Figure \ref{fig:Suppl_CRF-generic-model/architecture}.

\begin{table}
\resizebox{\textwidth}{!}{

\begin{tabular}{|c|c|c|}
\hline 
Parameter name & Set/range values & Best/optimal value\tabularnewline
\hline 
\hline 
$l_2$-norm regularization $\lambda$ & $\{10^{-3},10^{-2},10^{-1}\}$ & $10^{-2}$ \tabularnewline
\hline 
Batch size during training $|B|$ & $\{8,16,32,64,128\}$ & $64$ \tabularnewline
\hline 
Optimization algorithm & \{Adam\} & Adam \tabularnewline
\hline 
\end{tabular}
}

\caption{\label{tab:Suppl_CRF-hyperparameter-options}CRF hyperparameter options
(see Figure \ref{fig:Suppl_CRF-generic-model/architecture})}
\end{table}

\begin{table}
\resizebox{\textwidth}{!}{

\begin{tabular}{|c|c|c|}
\hline 
Parameter name & Set/range values & Best/optimal value\tabularnewline
\hline 
\hline 
Embedding layer (Blue block) dimension & $\{0,\lfloor d/2\rfloor,\lfloor d/3\rfloor,\lfloor d/4\rfloor\}$
where $d$ is input dimension & $\lfloor d/2\rfloor$ \tabularnewline
\hline 
Dropout probability $p_{dropout}$ & \{$0.15,0.35,0.5\}$ & $0.15$ \tabularnewline
\hline 
Embedding layer (Orange block) & $\{0,\lfloor D_{l}/2\rfloor,\lfloor D_{l}/3\rfloor,\lfloor D_{l}/4\rfloor\}$
where $D_{l}$ is input dimension from previous layer $l$ & $\lfloor D_{l}/3\rfloor$ \tabularnewline
\hline 
Non-linear function & $\{tanh,ReLU\}$ & $tanh$ \tabularnewline
\hline 
$l_2$-norm regularization $\lambda$ & $\{10^{-3},10^{-2},10^{-1}\}$ & $10^{-2}$ \tabularnewline
\hline 
Batch size during training $|B|$ & $\{8,16,32,64,128\}$ & $64$ \tabularnewline
\hline 
Optimization algorithm & \{Adam\} & Adam \tabularnewline
\hline 
\end{tabular}
}

\caption{\label{tab:Suppl_CRFNeural-hyperparameter-options}Neural CRF hyperparameter options
(see Figure \ref{fig:Suppl_CRF-generic-model/architecture})}
\end{table}

\subsubsection{CNN model}

\begin{figure}
\includegraphics[scale=0.031]{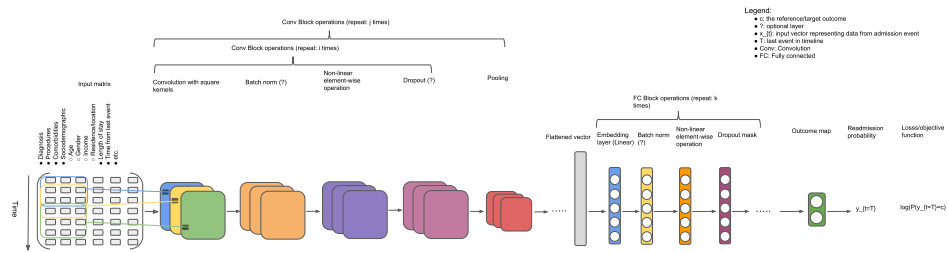}

\caption{\label{fig:Suppl_CNN-generic-model/architecture}CNN generic model/architecture}
\end{figure}

CNN model's hyperparameters configuration is reported in List \ref{alg:Suppl_List-of-CNN-hyperparameter-options}
along with the network architecture/design in Figure \ref{fig:Suppl_CNN-generic-model/architecture}.

\begin{algorithm}
\dirtree{%
.0 CNN hyperparameters configuration.
.1 Conv Block operations.
.2 Conv Block J.
.3 Conv Block I.
.4 Square kernel size \dotfill $\{\textcolor{blue}{3\times3}, 5\times5\}$.
.4 Batch norm \dotfill $\{\textcolor{blue}{True},False\}$.
.4 Non-linear function \dotfill $\{tanh,\textcolor{blue}{ReLU}\}$.
.4 Dropout \dotfill $\{\textcolor{blue}{0},0.15\}$.
.4 Starting number of kernels \dotfill $\{\textcolor{blue}{64},128,256\}$.
.4 Number of repeats for Conv Block I \dotfill $\{\textcolor{blue}{1},2,3\}$.
.3 Pooling \dotfill $\{AvgPool,\textcolor{blue}{MaxPool}\}$.
.3 Number of repeats for Conv Block J \dotfill $\{\textcolor{blue}{7},8\}$.
.1 FC Block operations.
.2 Embedding layer dimension \dotfill $\{\textcolor{blue}{\lfloor D_{l}/3\rfloor},\lfloor D_{l}/4\rfloor\,\lfloor D_{l}/5\rfloor\}$ where $D_{l}$ is the dimension of flattened feature vector from the last convolutional layer.
.2 Batch norm \dotfill $\{\textcolor{blue}{True},False\}$.
.2 Non-linear function \dotfill $\{tanh,\textcolor{blue}{ReLU}\}$.
.2 Dropout \dotfill $\{\textcolor{blue}{0},0.15,0.35,0.5\}$.
.2 Number of repeats for FC Block K \dotfill $\{\textcolor{blue}{1},2\}$.
.1 $l_2$-norm regularization $\lambda$ \dotfill $\{10^{-3},\textcolor{blue}{10^{-2}},10^{-1}\}$.
.1 Batch size during training $|B|$ \dotfill $\{8,\textcolor{blue}{16},32\}$.
.1 Optimization algorithm \dotfill $\{\textcolor{blue}{Adam}\}$.
}

\caption{CNN hyperparameter options \label{alg:Suppl_List-of-CNN-hyperparameter-options}
(see Figure \ref{fig:Suppl_CNN-generic-model/architecture}). Best parameters
are colored in \textcolor{blue}{blue}.}

\end{algorithm}

\subsubsection{CNN-Wide model}

\begin{figure}
\includegraphics[scale=0.045]{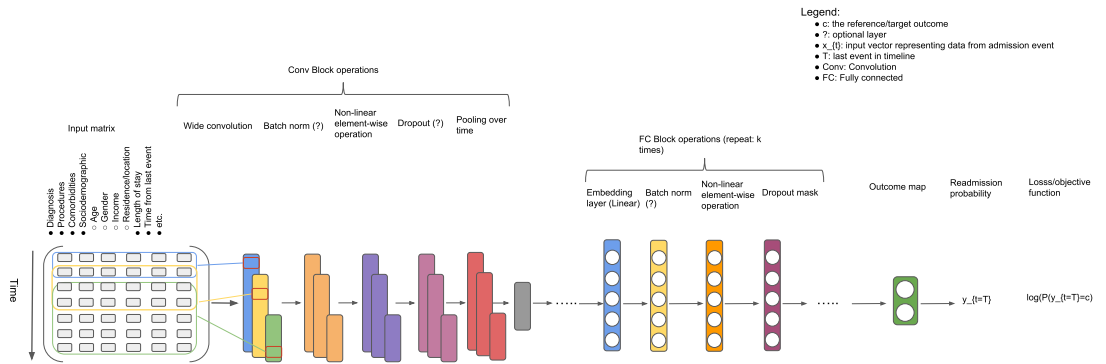}

\caption{\label{fig:Suppl_CNNWide-generic-model/architecture}CNN generic model/architecture}
\end{figure}

CNN-Wide model's hyperparameters configuration is reported in List
\ref{alg:Suppl_List-of-CNNWide-hyperparameter-options} along with the network
architecture/design in Figure \ref{fig:Suppl_CNNWide-generic-model/architecture}.

\begin{algorithm}
\dirtree{%
.0 CNN-Wide hyperparameters configuration.
.1 Conv Block operations.
.2 Rectangular kernel size \dotfill $\{\textcolor{blue}{2 \times d}, \textcolor{blue}{3 \times d},5 \times d\}$ where $d$ is the input dimension.
.2 Batch norm \dotfill $\{\textcolor{blue}{True},False\}$.
.2 Non-linear function \dotfill $\{\textcolor{blue}{tanh},ReLU\}$.
.2 Dropout \dotfill $\{\textcolor{blue}{0},0.15\}$.
.2 Number of kernels \dotfill $\{16,\textcolor{blue}{32},64,128\}$.
.2 Apply padding \dotfill $\{True,False\}$.
.2 Pooling \dotfill $\{AvgPool,\textcolor{blue}{MaxPool}\}$.
.2 Number of kernel types used \dotfill $\{\textcolor{blue}{2},3\}$.
.1 FC Block operations.
.2 Embedding layer dimension \dotfill $\{\textcolor{blue}{D_{l}},\lfloor D_{l}/2\rfloor,\lfloor D_{l}/3\rfloor,\lfloor D_{l}/4\rfloor\}$ where $D_{l}$ is the dimension of flattened feature vector from the last convolutional layer.
.2 Batch norm \dotfill $\{\textcolor{blue}{True},False\}$.
.2 Non-linear function \dotfill $\{\textcolor{blue}{tanh},ReLU\}$.
.2 Dropout \dotfill $\{\textcolor{blue}{0},0.15,0.35,0.5\}$.
.2 Number of repeats for FC Block K \dotfill $\{\textcolor{blue}{1},2\}$.
.1 $l_2$-norm regularization $\lambda$ \dotfill $\{10^{-3},\textcolor{blue}{10^{-2}},10^{-1}\}$.
.1 Batch size during training $|B|$ \dotfill $\{8,\textcolor{blue}{16},32\}$.
.1 Optimization algorithm \dotfill $\{\textcolor{blue}{Adam}\}$.
}

\caption{CNNWide hyperparameter options \label{alg:Suppl_List-of-CNNWide-hyperparameter-options}
(see Figure \ref{fig:Suppl_CNNWide-generic-model/architecture}). Best parameters
are colored in \textcolor{blue}{blue}.}
\end{algorithm}

\subsubsection{MLP model}

\begin{figure}[tbh]

\includegraphics[scale=0.08]{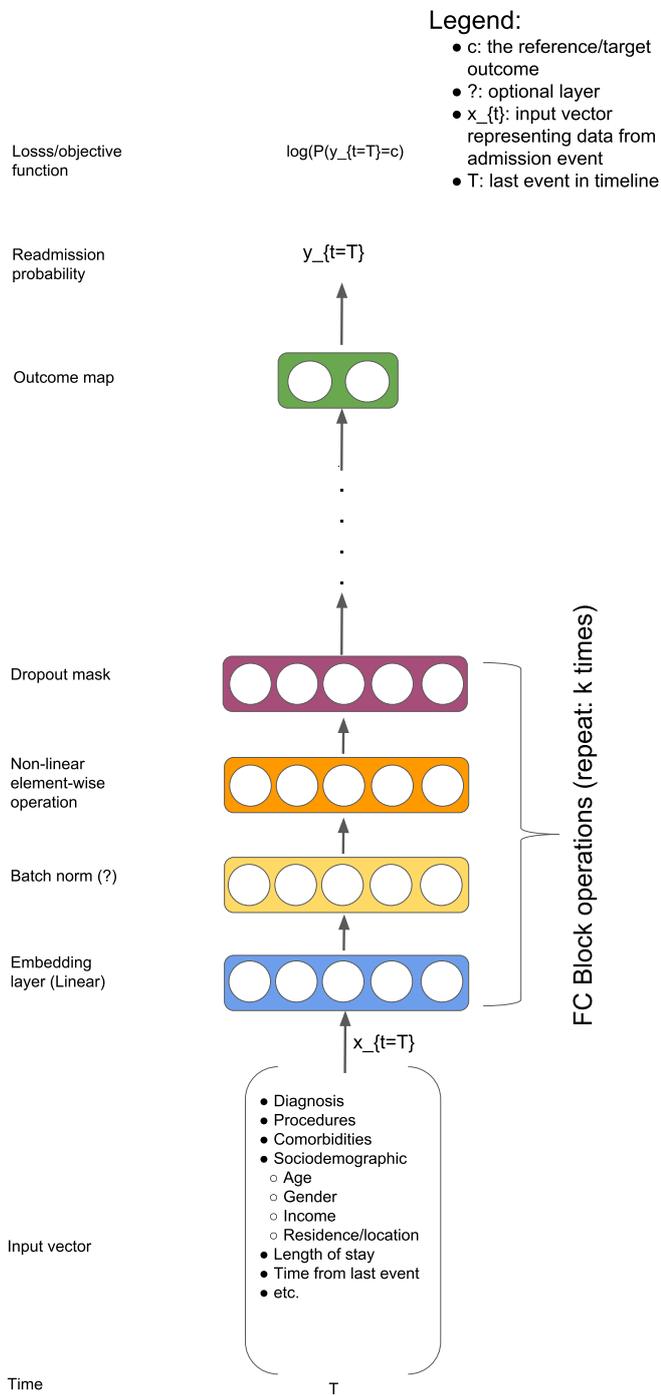}

\caption{\label{fig:Suppl_NN-generic-model/architecture}NN generic model/architecture}
\end{figure}

MLP model's hyperparameters configuration is reported in List \ref{alg:Suppl_List-of-NN-hyperparameter-options}
along with the network architecture/design in Figure \ref{fig:Suppl_NN-generic-model/architecture}.

\begin{algorithm}
\dirtree{%
.0 NN hyperparameters configuration.
.1 FC Block operations.
.2 Embedding layer dimension \dotfill $\{\lfloor D_{l}/2\rfloor,\lfloor D_{l}/3\rfloor,\textcolor{blue}{\lfloor D_{l}/4\rfloor}\}$ \newline where $D_{l}$ is the dimension of feature vector from the previous layer.
.2 Batch norm \dotfill $\{\textcolor{blue}{True},False\}$.
.2 Non-linear function \dotfill $\{tanh,\textcolor{blue}{ReLU}\}$.
.2 Dropout \dotfill $\{\textcolor{blue}{0},0.15,0.35,0.5\}$.
.2 Number of repeats for FC Block K \dotfill $\{1,\textcolor{blue}{2},3,4,5\}$.
.1 $l_2$-norm regularization $\lambda$ \dotfill $\{10^{-3},\textcolor{blue}{10^{-2}},10^{-1}\}$.
.1 Batch size during training $|B|$ \dotfill $\{32,64,\textcolor{blue}{128}\}$.
.1 Optimization algorithm \dotfill $\{\textcolor{blue}{Adam}\}$.
}

\caption{NN hyperparameter options \label{alg:Suppl_List-of-NN-hyperparameter-options}
(see Figure \ref{fig:Suppl_NN-generic-model/architecture}). Best parameters
are colored in \textcolor{blue}{blue}.}
\end{algorithm}

% \begin{figure}[tbh]

% \FloatBarrier

\subsubsection{Logistic regression}

Logistic regression models' hyperparameters options are reported
in Tables \ref{tab:Suppl_logitl1-hyperparameter-options} and \ref{tab:Suppl_logitl2-hyperparameter-options} respectively.

\begin{table}
\resizebox{\textwidth}{!}{
\begin{tabular}{|c|c|c|}
\hline 
Parameter name  & Set/range values & Best/optimal value\tabularnewline
\hline 
\hline 
$l_{1}$-norm regularization $\lambda$ & $\{10^{-3},10^{-2},10^{-1}\}$ & $10^{-1}$\tabularnewline
\hline 
Optimization algorithm & \{Liblinear, Saga\} & Liblinear/Saga \tabularnewline
\hline 
Weighting scheme (i.e. weighting training examples) & \{Balanced, None\} & Balanced\tabularnewline
\hline
\end{tabular}
}
\caption{\label{tab:Suppl_logitl1-hyperparameter-options}Logistic regression with $l_1$-norm regularization (LASSO)}
\end{table}

\begin{table}
\resizebox{\textwidth}{!}{

\begin{tabular}{|c|c|c|}
\hline 
Parameter name  & Set/range values & Best/optimal value\tabularnewline
\hline 
\hline 
$l_2$/$l_{1}$-norm regularization $\lambda$ & $\{10^{-3},10^{-2},10^{-1},1\}$ & $10^{-1}$\tabularnewline
\hline 
Optimization algorithm & \{Liblinear, Saga\} & Liblinear/Saga \tabularnewline
\hline 
Weighting scheme (i.e. weighting training examples) & \{Balanced, None\} & Balanced\tabularnewline
\hline
\end{tabular}

}

\caption{\label{tab:Suppl_logitl2-hyperparameter-options}Logistic regression with $l_2$-norm regularization}
\end{table}
\subsection{Feature importance}
\subsubsection{Logistic regression}
\begin{figure}[tbh]
\includegraphics[scale=0.24]{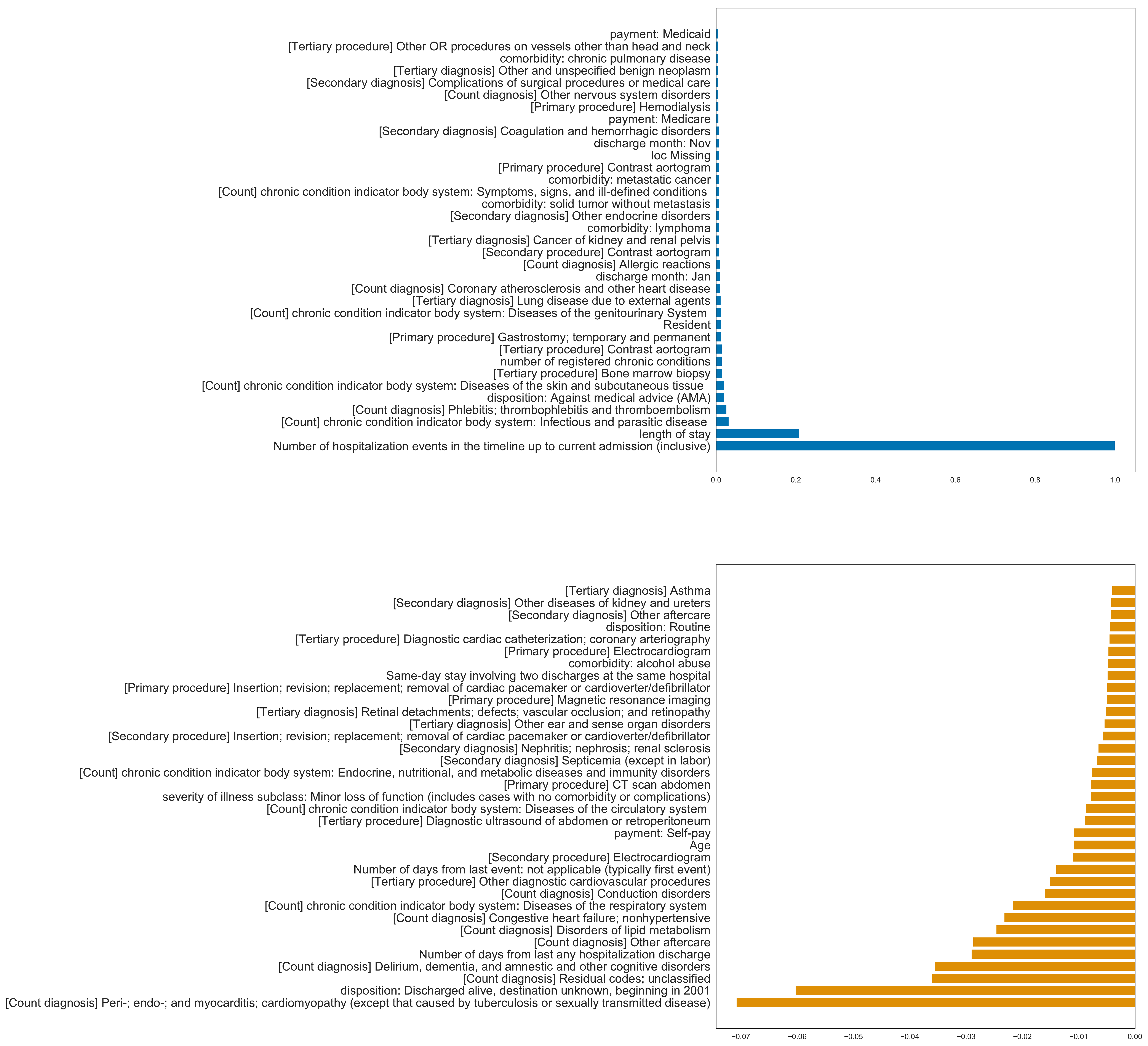}
\caption{\label{fig:Suppl_Top-features-logit}Top-35 features (normalized) in LASSO models contributing to the increase and decrease of log-odds of readmission}
\end{figure}

The analysis of feature importance is
reported in Figure \ref{fig:Suppl_Top-features-logit}, which shows the normalized
coefficients of the trained LASSO models averaged across all folds. 
\subsubsection{RNNCRF model}
\begin{figure}[tbh]
\includegraphics[scale=0.25]{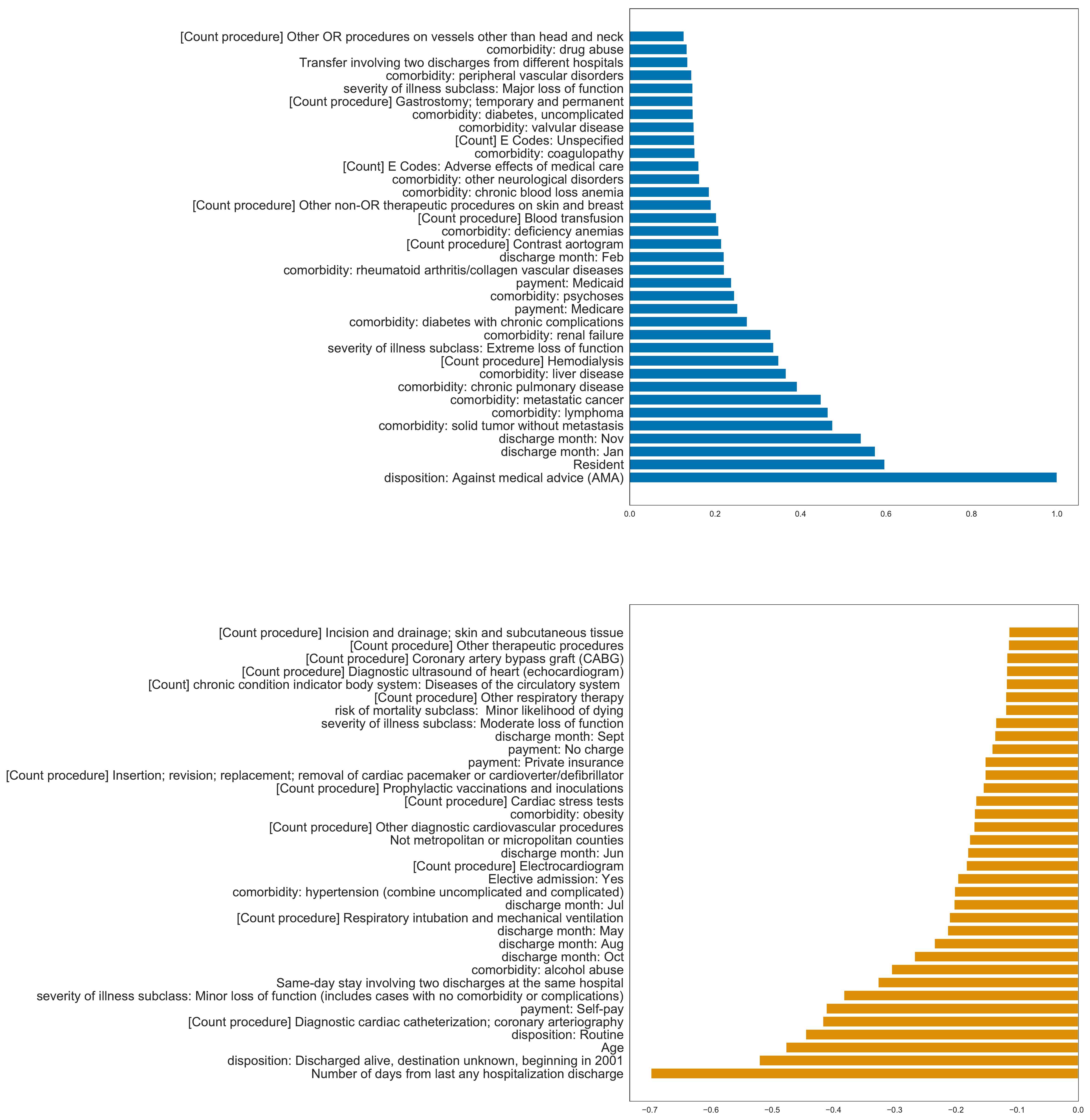}
\caption{\label{fig:Suppl_Top-features-diffprob}Top-35 features (normalized) in RNNCRF models using \textit{diff\_prob} metric}
\end{figure}

\begin{figure}[tbh]
\includegraphics[scale=0.25]{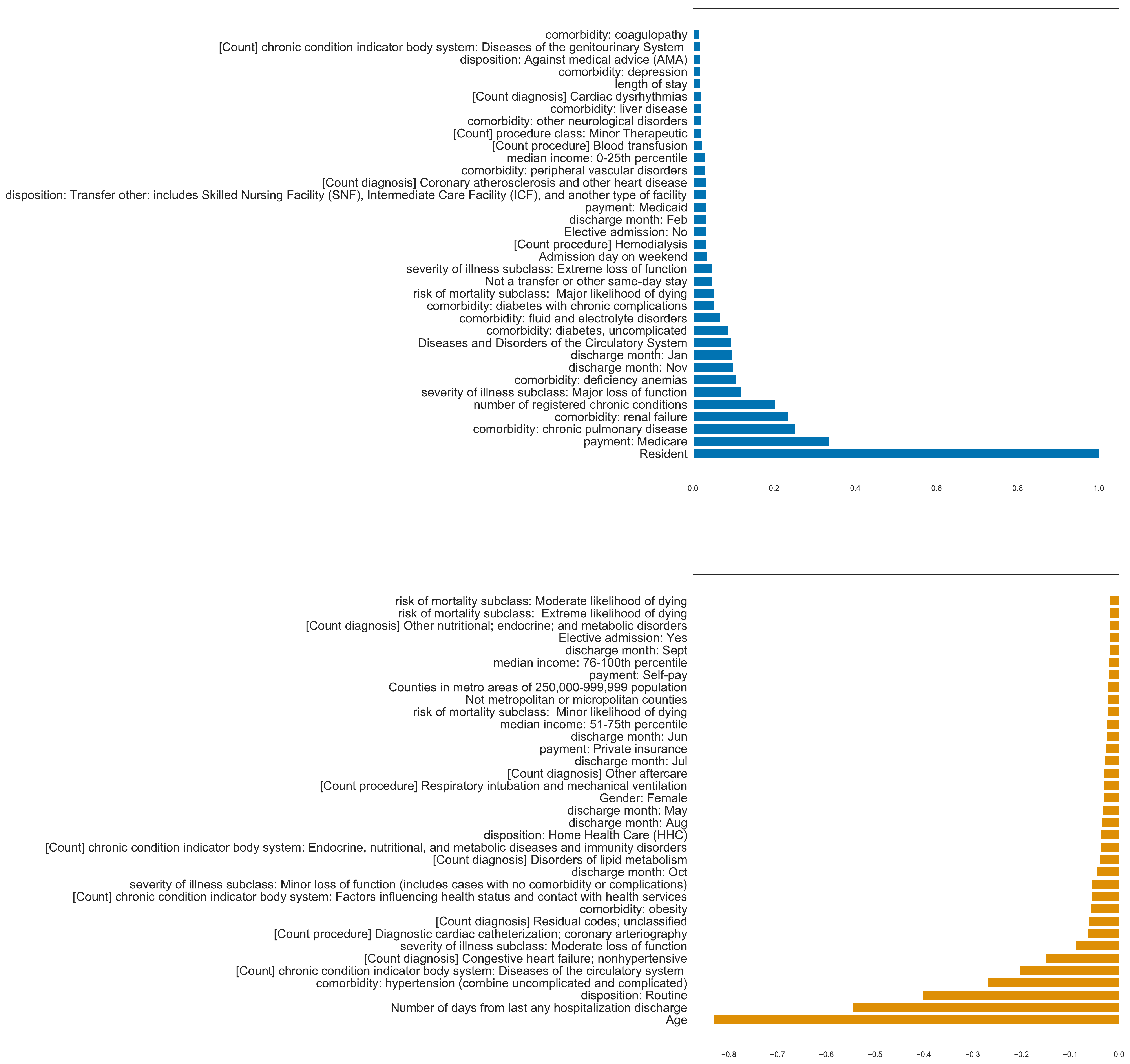}
\caption{\label{fig:Suppl_Top-features-diffprobweighted}Top-35 features (normalized) in RNNCRF models using \textit{diff\_prob\_weighted} metric}
\end{figure}

\begin{figure}[tbh]
\includegraphics[scale=0.25]{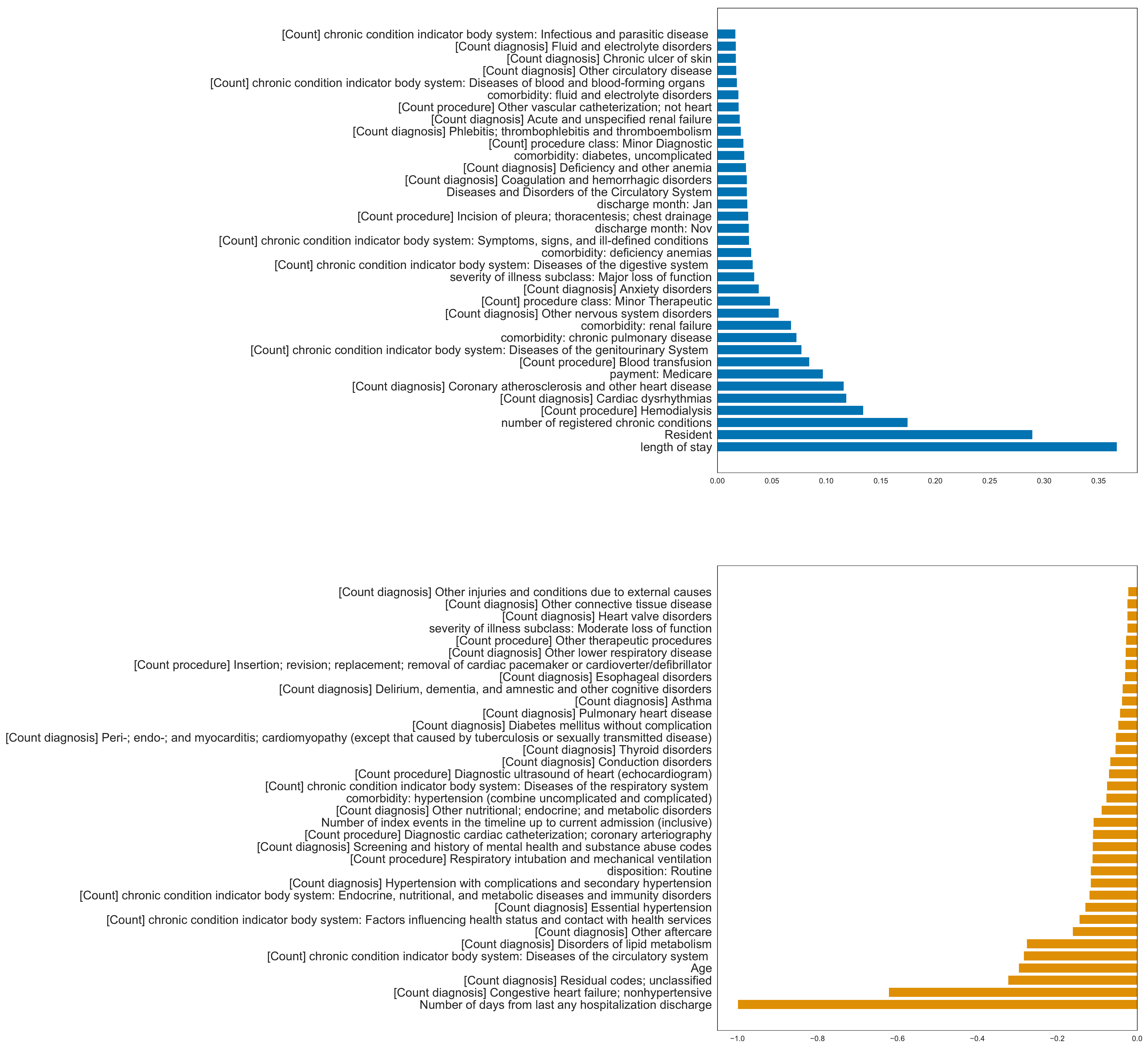}
\caption{\label{fig:Suppl_Top-features-ratiodiffprobweighted}Top-35 features (normalized) in RNNCRF models using \textit{ratio\_diff\_prob\_weighted} metric}
\end{figure}

For the best neural model (RNNCRF), we report the analysis of feature importance according to an approach previously reported
in \cite{Avati2017}. In short, we iterated over all features attached to the last
HF event, and computed the probability of readmission with a feature present or absent. The difference between both probabilities allowed us to quantify a feature's importance across the five folds (we call this metric \textit{diff\_prob} see Fig. \ref{fig:Suppl_Top-features-diffprob}). Additionally, we computed another variation of the same metric by incorporating the percentage of occurrence of each feature (i.e. when the feature is present) in the computation. In other words, we weighted the computed differences by the percentage of time each feature was present in the dataset (referred to \textit{diff\_prob\_weighted}, see Fig. \ref{fig:Suppl_Top-features-diffprobweighted}). A third variation, is computing a ratio (for every feature) dividing the average difference in probability (\textit{diff\_prob}) by the average value of the feature when it was present and again weighted by the percentage of occurrence of the feature (\textit{ratio\_diff\_prob\_weighted}, see Fig. \ref{fig:Suppl_Top-features-ratiodiffprobweighted}).

\end{document}